\documentclass[10pt,twocolumn,letterpaper]{article}

\usepackage{cvpr}
\usepackage{times}
\usepackage{epsfig}
\usepackage{graphicx}
\usepackage{amsmath}
\usepackage{amssymb}

\usepackage{enumitem}
\usepackage{scrextend}
\usepackage{subcaption}
\usepackage[dvipsnames]{xcolor}
\usepackage{algorithm}
\usepackage[noend]{algpseudocode}
\usepackage{ctable} % for \specialrule command

\definecolor{citecolor}{RGB}{119,185,0}
\usepackage[pagebackref=true,breaklinks=true,letterpaper=true,colorlinks,citecolor=citecolor,bookmarks=false]{hyperref}

\cvprfinalcopy % *** Uncomment this line for the final submission

\def\cvprPaperID{2838}

\ifcvprfinal\fi
\begin{document}

\title{Instance-Aware, Context-Focused, and Memory-Efficient\\
Weakly Supervised Object Detection}

\author{
Zhongzheng Ren$^{1,2}$\footnotemark[1]
\quad Zhiding Yu$^{2}$
\quad Xiaodong Yang$^{2}$\footnotemark[1]
\quad Ming-Yu Liu$^{2}$\\
\quad Yong Jae Lee$^{3}$
\quad Alexander G.  Schwing$^{1}$
\quad Jan Kautz$^{2}$\\
$^{1}$University of Illinois at Urbana-Champaign\quad $^{2}$NVIDIA\quad $^{3}$University of California, Davis
}

\maketitle

\begin{abstract}
\vspace{-0.1in}
Weakly supervised learning has emerged as a compelling tool for object detection by reducing the need for strong supervision during training. However, major challenges remain: (1) differentiation of object instances can be ambiguous; (2) detectors tend to focus on discriminative parts rather than entire objects; (3) without ground truth, object proposals have to be redundant for high recalls, causing significant memory consumption. Addressing these challenges is difficult, as it often requires to eliminate uncertainties and trivial solutions. To target these issues we develop an instance-aware and context-focused unified framework. It employs an instance-aware self-training algorithm and a learnable Concrete DropBlock while devising a memory-efficient sequential batch back-propagation. Our proposed method achieves state-of-the-art results on COCO (12.1\% AP, 24.8\% $AP_{50}$), VOC 2007 (54.9\% AP),  and VOC 2012 (52.1\% AP), improving baselines by great margins. In addition, the proposed method is the first to benchmark ResNet based models and weakly supervised video object detection. 
Code, models, and more details will be made available at:~\url{https://github.com/NVlabs/wetectron}.
\end{abstract}
\vspace{-1em}
\footnotetext[1]{Work partially done at NVIDIA.}

% !TEX root = ./main.tex
\section{Introduction}
\label{sec:intro}
Recent works on object detection~\cite{he2017maskrcnn, ren16faster, yolo, cornernet} have achieved impressive results. However, the training process often requires strong supervision in terms of precise bounding boxes. Obtaining such annotations at a large scale can be costly, time-consuming, or even infeasible. This motivates weakly supervised object detection (WSOD) methods~\cite{Bilen16, tang2017multiple, KantorovOCL16} where detectors are trained with weaker forms of supervision such as image-level category labels. These works typically formulate WSOD as a multiple instance learning task, treating the set of object proposals in each image as a bag. The selection of proposals that truly cover objects is modeled using learnable latent variables. 

\begin{figure}[t]
\centering
\includegraphics[width=0.48\textwidth]{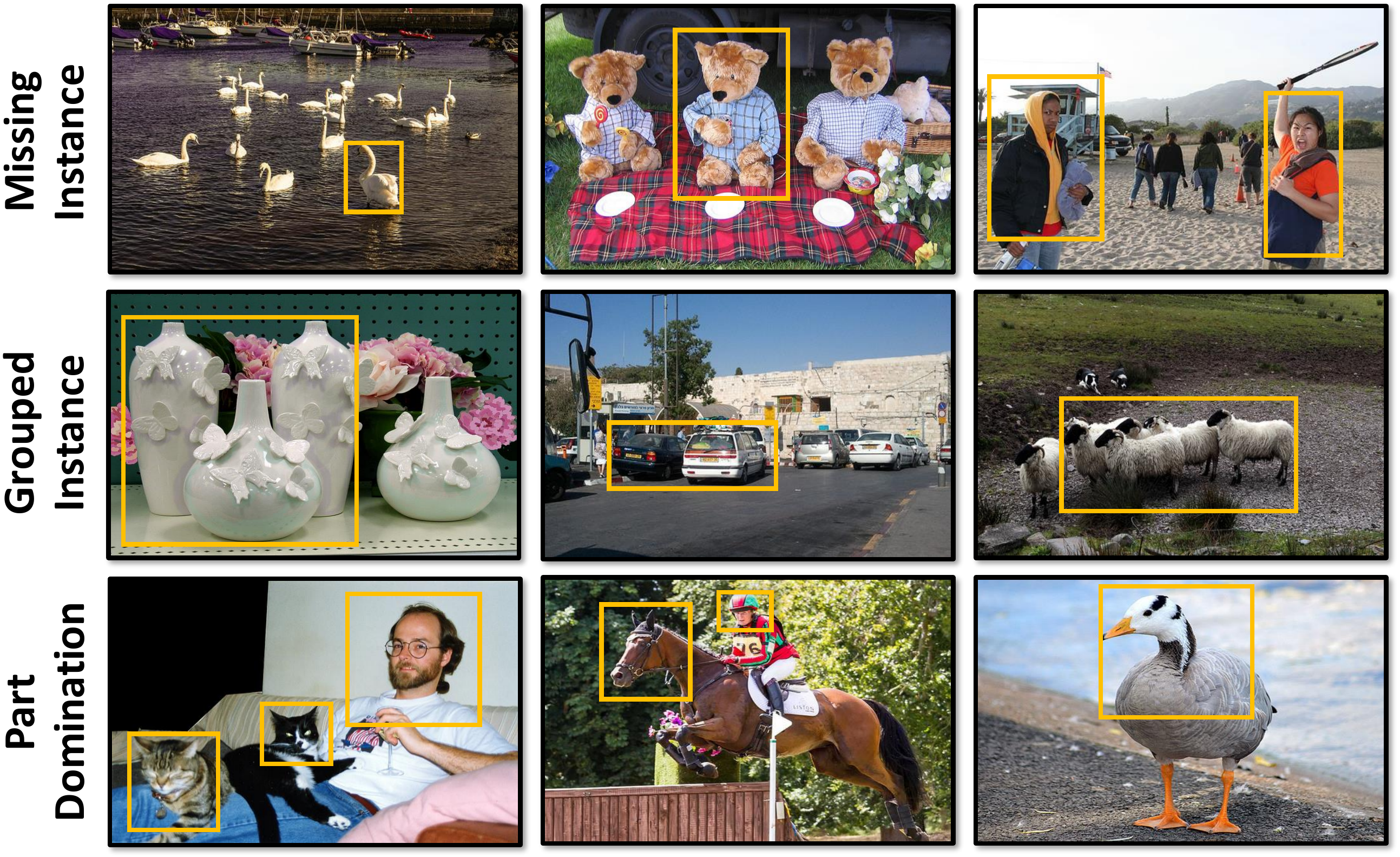}
\caption{Typical WSOD issues: (1) \textbf{Instance Ambiguity:} missing less salient objects (top) or failing to differentiate clustered instances (middle); (2) \textbf{Part Domination:} focusing on most discriminative object parts (bottom). }
\label{fig:teaser}
\vspace{-1.5em}
\end{figure}

While alleviating the need for precise annotations, existing weakly supervised object detection methods~\cite{Bilen16, tang2017multiple, c-mil, singh-cvpr2019, Zeng_2019_ICCV} often face three major challenges due to the under-determined and ill-posed nature, as demonstrated in Fig.~\ref{fig:teaser}: 

\noindent\textbf{{(1) Instance Ambiguity.}} This arguably the biggest challenge which subsumes two common types of issues: 
(a) \textit{Missing  Instances:} Less salient objects in the background with rare poses and smaller scales are often ignored (top row in Fig.~\ref{fig:teaser}). 
(b) \textit{Grouped Instances:} Multiple instances of the same category are grouped into a single bounding box when spatially adjacent (middle row in Fig.~\ref{fig:teaser}).
Both issues are caused by bigger or more salient boxes receiving higher scores than smaller or less salient ones. 

\noindent\textbf{{(2) Part Domination.}}  Predictions tend to be dominated by the most discriminative parts of an object (Fig.~\ref{fig:teaser} bottom). This issue is particularly pronounced for classes with big intra-class difference. For example, on classes such as animals and people, the model often turns into a `face detector' as faces are the most consistent appearance signal.

\noindent\textbf{{(3) Memory Consumption.}} Existing proposal generation methods~\cite{ss, eb} often produce dense %number of 
proposals. Without ground-truth localization, maintaining a large number of proposals is necessary to achieve a reasonable recall rate and good performance. This requires a lot of memory, especially for video object detection. Due to the large number of proposals, most memory is consumed in the intermediate layers after ROI-Pooling. 

To address the above three challenges, we propose a unified weakly supervised learning framework that is instance-aware and context-focused. The proposed method tackles \textbf{{Instance Ambiguity}} by introducing an advanced self-training algorithm where instance-level pseudo ground-truth, in forms of category labels and regression targets are computed by considering more instance-associative spatial diversification constraints (Sec.~\ref{sec:refine}). The proposed method also addresses \textbf{Part Domination} by introducing a parametric spatial dropout termed `Concrete DropBlock.' This module is learned end-to-end to adversarially maximize the detection objective, thus encouraging the whole framework to consider context rather than focusing on the most discriminative parts (Sec.~\ref{sec:dropout}). Finally, to alleviate the issue of \textbf{{Memory Consumption}}, our method adopts a sequential batch back-propagation algorithm which processes data in batches at the most memory-heavy stage. This permits the assess to larger deep models such as ResNet~\cite{resnet} in WSOD, as well as the exploration of weakly supervised video object detection (Sec.~\ref{sec:step-bp}).

Tackling the aforementioned three challenges via our proposed framework leads to state-of-the-art performance on several popular datasets, including COCO~\cite{coco}, VOC 2007 and 2012~\cite{pascal}. The effectiveness and robustness of each proposed module is demonstrated in detailed ablation studies, and further verified through qualitative results. Finally, we conduct additional experiments on videos and give the first benchmark for weakly supervised video object detection on ImageNet VID~\cite{imagenet}.

% !TEX root = ./main.tex

\section{Related work}
\label{related}

\paragraph{Weakly supervised object detection (WSOD).}
Object detection is one of the most fundamental problems in computer vision. Recent supervised methods~\cite{rcnn, fastrcnn, ren16faster, he2017maskrcnn, yolo, ssd, cornernet} have shown great performance in terms of both accuracy and speed. For WSOD, most methods formulate a multiple instance learning problem where input images contain a bag of instances (object proposals). The model is trained with a classification loss to select the most confident positive proposals. Modifications \wrt initialization~\cite{song14slsvm, siva_iccv13}, regularization~\cite{CinbisVS14,Bilen14b, WangZYB15}, and representations~\cite{CinbisVS14,BilenPT15, LiHLW016} have been shown to improve results. For instance, Bilen and Vedaldi~\cite{Bilen16} proposed an end-to-end trainable architecture for this task. Follow-up works further improve by leveraging spatial relations~\cite{tang2017multiple, tang2018pcl, KantorovOCL16}, better optimization~\cite{zigzag, JieWJFL17, Arun_2019, c-mil}, and multitasking with weakly supervised segmentation~\cite{Ge_2018_CVPR, Shen_2019_CVPR, Gao_2019_ICCV, singh-cvpr2019}.  

\vspace{-1em}
\paragraph{Self-training for WSOD.} 
Among the above directions, self-training~\cite{zou2018unsupervised,Zou_2019_ICCV} has been demonstrated to be seminal. Self-training uses instance-level pseudo labels to augment training and can be implemented in an \textbf{offline} manner~\cite{Zhang_2018_CVPR, krishna-cvpr2016, LiHLW016, Zhang_2018_CVPR}: a WSOD model is first trained using any of the methods discussed above; then the confident predictions  are used as pseudo-labels to train a final supervised detector. This iterative knowledge distillation procedure is beneficial since the additional supervised models learn form less noisy data and usually have better architectures for which training is  time-consuming. A number of works~\cite{tang2017multiple, tang2018pcl, c-mil, Gao_2019_ICCV, Zeng_2019_ICCV, TangWWYLHY18} studied end-to-end implementations of self-training: WSOD models compute and use pseudo labels simultaneously during training, which is commonly referred to as an \textbf{online} solution.
However, these methods typically only consider the most confident predictions for pseudo-labels. Hence they tend to have overfitting issues with difficult parts and instances ignored.

\vspace{-1em}
\paragraph{Spatial dropout.}
To address the above issue, an effective regularization strategy is to drop parts of spatial feature maps during training. Variants of spatial-dropout have been widely designed for supervised tasks such as classification~\cite{dropblock}, object detection~\cite{a-fast-rcnn}, and human joints localization~\cite{TompsonGJLB15}. Similar approaches have also been applied in weakly supervised tasks for better localization in detection~\cite{singh-iccv2017} and semantic segmentation~\cite{WeiFLCZY17}. However, these methods are non-parametric and cannot adapt to different datasets in a data-driven manner. As a further improvement, Kingma \etal~\cite{varitional-db} designed variational dropout where the dropout rates are learned during training. Wang \etal~\cite{a-fast-rcnn} proposed a parametric but non-differentiable spatial-dropout trained with REINFORCE~\cite{reinforce}. In contrast, the proposed `Concrete DropBlock' module has a parametric and differentiable structured novel form.

\vspace{-1em}
\paragraph{\textbf{Memory efficient back-propagation.}}
Memory has always been a concern since deeper models~\cite{resnet,vgg} and larger batch size~\cite{Peng_2018_CVPR} often tend to yield better results. One way to alleviate this concern is to trade computation time for memory consumption by modifying the back-propagation (BP) algorithm~\cite{Rumelhart}. A suitable technique~\cite{Kokkinos17,PleissCHLMW17, ChenXZG16} is to not store some intermediate deep net representations during forward-propagation. One can recover those by injecting small forward passes during  back-propagation. Hence, the one-stage back-propagation is divided into several step-wise processes. However, this method cannot be directly applied to our model where a few intermediate layers consume most of the memory. To address it, we suggest a batch operation for the memory-heavy intermediate layers. 

% !TEX root = ./main.tex
\section{Background}
\label{bcakground}
Bilen and Vedaldi~\cite{Bilen16} are among the first to develop an end-to-end deep WSOD framework based on the idea of multiple instance learning. Specifically, given an input image $I$ and the corresponding set of pre-computed~\cite{ss, eb} proposals  $R$, an ImageNet~\cite{imagenet} pre-trained neural network is used to produce classification logits $f_w(c,r)\in\mathbb{R}$ and detection logits $g_w(c,r)\in\mathbb{R}$ for every object category $c\in C$ and for every region $r\in R$. The vector $w$ subsumes all trainable parameters. Two score matrices, \ie, $s(c|r)$ of a region $r$ being classified as category $c$, and $s(r|c)$ of detecting region $r$ for category $c$  are obtained through 
\begin{equation}
% \resizebox{\columnwidth}{!}{
\footnotesize{
s_w(c|r) = \frac{\exp f_w(c,r)}{\sum_{c\in C}\exp f_w(c,r)}, \text{~and~}
s_w(r|c) = \frac{\exp g_w(c,r)}{\sum_{r\in R}\exp g_w(c,r)}.
}
\label{softmax}
\end{equation} 
The final score $s_w(c,r)$ for assigning category $c$ to region $r$ is computed via an element-wise product: $s_w(c,r) =  s_w(c|r)s_w(r|c) \in [0,1]$. During training, $s_w(c,r)$ is summed for all regions $r \in R$ to obtain the image evidence $\phi_w(c)=\sum_{r\in R}s_w(c,r)$. The loss is then computed via:
\begin{equation}
\footnotesize{
\mathcal{L}_\text{img}(w) = -\sum_{c\in C} y(c)\log\phi_w(c), 
\label{loss_img}}
\end{equation}
where $y(c)\in\{0,1\}$ is the ground truth (GT) class label indicating image-level existence of category $c$. For inference, $s_w(c,r)$ is used for prediction followed by standard non-maximum suppression (NMS) and thresholding.

To integrate online self-training, the region score $s_w(c,r)$ is often used as teacher to generate instance-level pseudo category label $\hat y(c,r) \in \{0,1\}$ for every region $r\in R$~\cite{tang2018pcl, c-mil, Gao_2019_ICCV, Zeng_2019_ICCV, TangWWYLHY18}. This is done by treating the top-scoring region and its highly-overlapped neighbors as the positive examples for class $c$. The extra student layer is then trained for region classification via:
\begin{equation}
\footnotesize{
\mathcal{L}_\text{roi}( w) = - \frac{1}{|R|} \sum_{c \in C} \hat{y}(c,r) \log {\hat s_w(c|r) },
\label{loss_roi}}
\end{equation}
where $\hat s_w(c|r) $ is the output of this layer. During testing, the student prediction $\hat s_w(c|r) $ will be used rather than $s_w(c,r)$. We build upon this formulation and develop two additional novel modules as described subsequently.

% !TEX root = ./main.tex
\section{Approach}
\label{sec:approach}
\begin{figure}[t]
\centering
\includegraphics[width=0.48\textwidth]{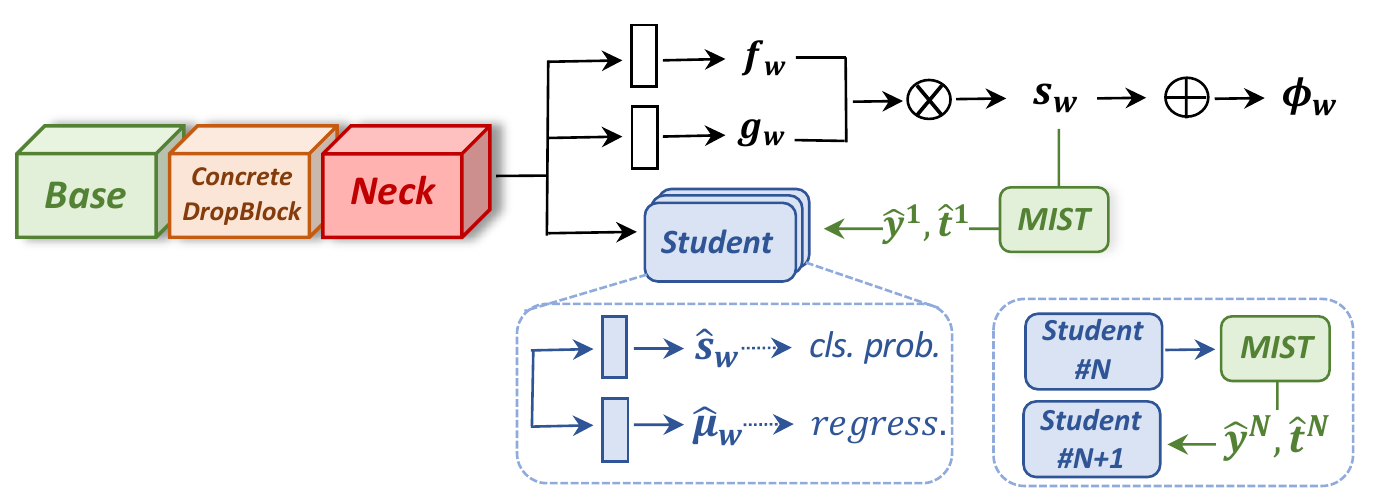}
\caption{The overall framework. ROI-Pooling and the operations in Eq.~\eqref{softmax} are abstracted away for readability.}
\label{fig:head}
\vspace{-1em}
\end{figure}

Image-level labels are an effective form of supervision to mine for common patterns across images. Yet inexact supervision often causes localization ambiguity. To address the mentioned three challenges caused by this ambiguity, we develop the instance-aware and context-focused framework outlined in Fig.~\ref{fig:head}. It contains a novel online self-training algorithm with ROI regression to reduce instance ambiguity and better leverage the self-training supervision (Sec.~\ref{sec:refine}). It also reduces part-domination for classes with large intra-class variance via a novel end-to-end learnable `Concrete DropBlock' (Sec.~\ref{sec:dropout}), and it is more memory friendly (Sec.~\ref{sec:step-bp}). 

\subsection{Multiple instance self-training (MIST)}
\label{sec:refine}

With online or offline generated pseudo-labels~\cite{tang2018pcl, krishna-cvpr2016, Zhang_2018_CVPR}, self-training helps to eliminate localization ambiguities, benefiting mainly from two aspects: (1) Pseudo-labels permit to model proposal-level supervision and inter-proposal relations; (2) Self-training can be broadly regarded as a teacher-student distillation process which has been found helpful to improve the student's representation. We take the following dimensions into account when designing our framework: 

\noindent\textbf{Instance-associative:} Object detection is often `instance-associative': highly overlapping proposals should be assigned similar labels. Most self-training methods for WSOD ignore this  and instead treat  proposals  independently. Instead, we impose explicit instance-associative constraints into pseudo box generation. 

\noindent\textbf{Representativeness:} The score of each proposal in general is a good proxy for its representativeness. It is not perfect, especially in the beginning there is a tendency to focus on object parts. However, the score provides a high recall for being at least located on correct objects. 

\noindent\textbf{Spatial-diversity:} Imposing spatial diversity to the selected pseudo-labels can be a useful self-training inductive bias. It promotes better coverage on difficult (\eg, rare appearance, poses, or occluded) objects, and higher recall for multiple instances (\eg, diverse scales and sizes).

The above constraints and criteria motivate  a novel algorithm  to generate diverse yet representative pseudo boxes which are  instance-associative. The details are provided in Alg.~\ref{alg:inst-or}. Specifically, we first sort all the scores across the set $R$ for each class $c$ that appears in the category-label. We then pick the top $p$ percent of the ranked regions to form an initial candidate pool $R'(c)$. Note that the size of the candidate pool $R'(c)$, \ie, $|R'(c)|$ is image-adaptive and content-dependent by being proportional to $|R|$. Intuitively, $|R|$ is a meaningful prior for the overall objectness of an input image. A diverse set of high-scoring non-overlapping regions are then picked from $R'(c)$ as the pseudo boxes $\hat R(c)$ using non-maximum suppression. Even though being simple, this effective algorithm leads to significant performance improvements as shown in Sec.~\ref{sec:exp}. 

\begin{algorithm}[t]
\footnotesize
\caption{Multiple Instance Self-Training}
\begin{algorithmic}[1]
      \Require{Image $I$, class label $y$, proposals $R$, threshold $\tau$, percentage $p$}
      \Ensure{Pseudo boxes $\hat{R}^1$}
        \State Feed $I$ into model; get ROI scores $s$
        \For{ground-truth class $c$}
            \State $R(c)_{sorted} \leftarrow ~\text{SORT}(s(c, *)) $  \hfill//sort ROIs by scores of class $c$
            \State $R'(c) \leftarrow $  top $p$ percent of $R(c)_{sorted}$
            \State  $\hat{R}(c) \leftarrow r'_0 $    \hfill//~ save first region (top-scoring) $ r'_0 \in R'$
            \For{$i$ ~in~ \{2~...~$|R'(c)|\}$}   \hfill//~start from the second highest
                \State $\text{APPEND}(\hat R(c), r'_i)$ \textbf{if}~ $\text{IoU}(r'_i, \hat r_j) < \tau, \forall \hat r_j \in R' (c): j < i$ 
            \EndFor
        \EndFor
        \State \textbf{return  $\hat{R}(c)$}
\end{algorithmic}
\label{alg:inst-or}
\end{algorithm}

\vspace{-1em}
\paragraph{Self-training with regression.} 
Bounding box regression is another module that plays an important role in supervised object detection but is missing in online self-training methods. To close the gap, we encapsulate a classification layer and a regression layer into `student blocks' as shown via blue boxes in Fig.~\ref{fig:head}.  We jointly optimize them using pseudo-labels $\hat R$. The predicted bounding boxes from the regression layer are referred to via $\mu_w(r)$ for all regions $ r \in R$. For each region $r$, if it is highly overlapping with a pseudo-box $\hat r \in \hat R$ for ground-truth class $c$, we  generate the regression target $\hat t(r)$ by using the coordinates of $\hat r$ and by marking the classification label $\hat y(c,r)=1$. The complete region-level loss for training the student block is:
\begin{align}
\footnotesize
\begin{split}
\mathcal{L}_\text{roi}(w) = \frac{1}{|R|}  \sum_{r\in R}  \lambda_r ( &
 \mathcal{L}_{\text{smooth-L1}} (\hat{t}(r), \mu_w(r))  \\[-6pt]
& -\frac{1}{|C|} \sum_{c \in C} \hat{y}(c,r) \log {\hat s_w(c|r)} ),
\label{eq:new_roi}
\end{split}
\end{align} \\[-10pt]
where $\mathcal{L}_{\text{smooth-L1}}$ is the Smooth-L1 objective  used in~\cite{fastrcnn} and $\lambda_r$ is a scalar per-region weight  used in~\cite{tang2017multiple}. 

In practice, conflicts happen when we force the $\hat{y}(\cdot,r)$ to be a one-hot vector since the same region can be chosen to be positive for different ground-truth classes, especially in the early stages of training. Our solution is to use that class for pseudo-label $\hat r$ which has a higher predicted score $s(c, \hat r)$. In addition, the obtained pseudo-labels and the proposals are inevitably  noisy. Imposing bounding box regression is able to correctly learn from the noisy labels by capturing the most consistent patterns among them, and refining the noisy proposal coordinates accordingly. We empirically verify in Sec.~\ref{exp:ablation} that bounding box regression improves both robustness and generalization. 

\vspace{-0.5em}
\paragraph{Self-ensembling.}
We follow~\cite{tang2017multiple, tang2018pcl} to stack multiple student blocks to improve  performance. As shown in Fig.~\ref{fig:head}, the first pseudo-label $\hat R^1$ is generated from the teacher branch, and then the student block $N$  generates  pseudo-label $\hat R^N$ for the next student block $N+1$. This technique is similar to the self-ensembling method~\cite{Laine2017}. 

\subsection{Concrete DropBlock}
\label{sec:dropout}

\begin{figure}[t]
\centering
\includegraphics[width=0.48\textwidth]{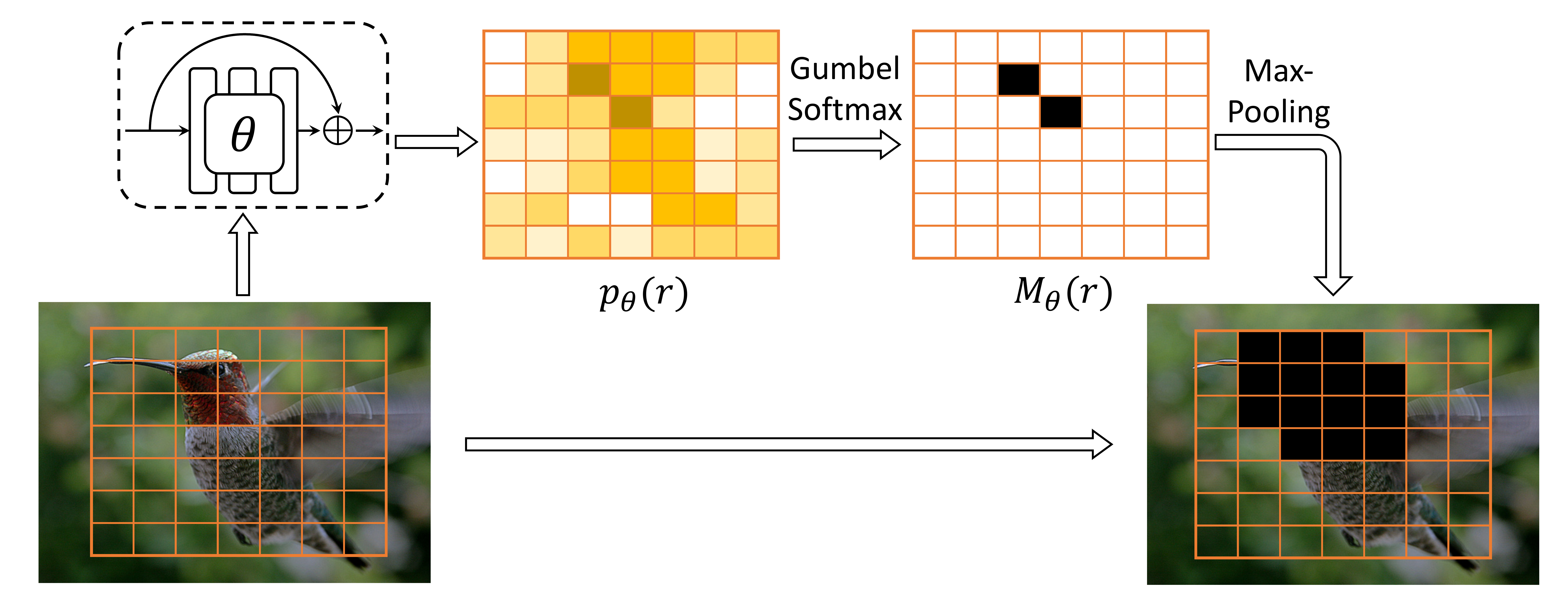}
\vspace{-2em}
\caption{Illustration of the Concrete DropBlock idea. Discriminative parts such as head are zeroed out.} 
\label{fig:dropblock}
\vspace{-1em}
\end{figure}

\begin{figure*}[t]
\centering
\begin{minipage}{.28\textwidth}
\centering
\includegraphics[width=\linewidth]{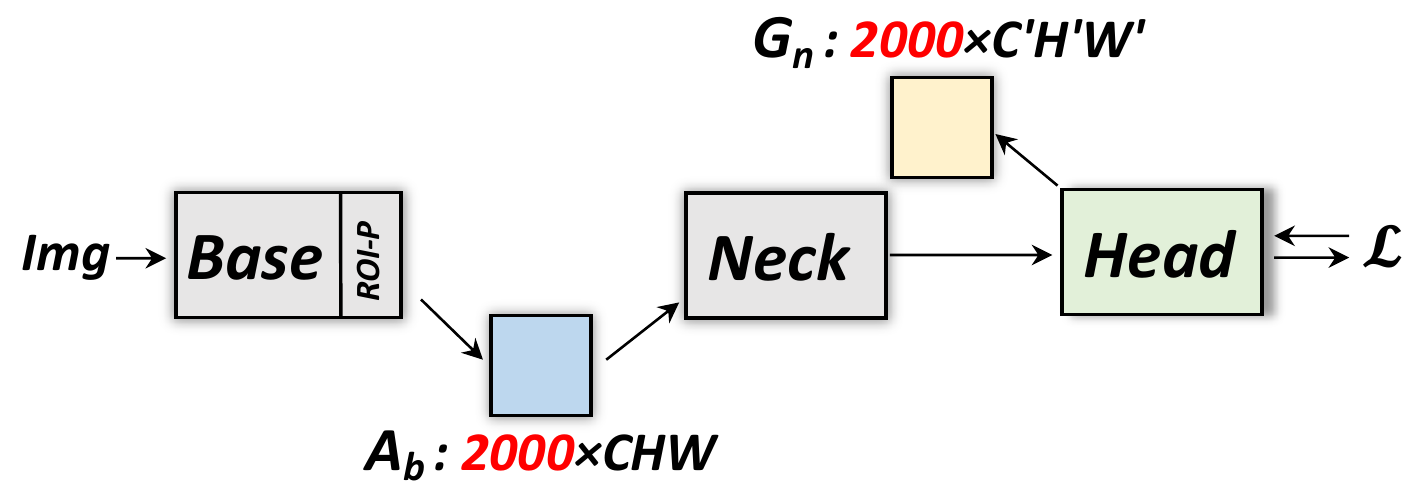}
\vspace{-2.5em}
\captionsetup{labelformat=empty}
\caption{(a) Forward and back-prop to \\update `Head'. $A_b, G_n$ saved.}
\end{minipage}
\begin{minipage}{.48\textwidth}
\centering
\includegraphics[width=\linewidth]{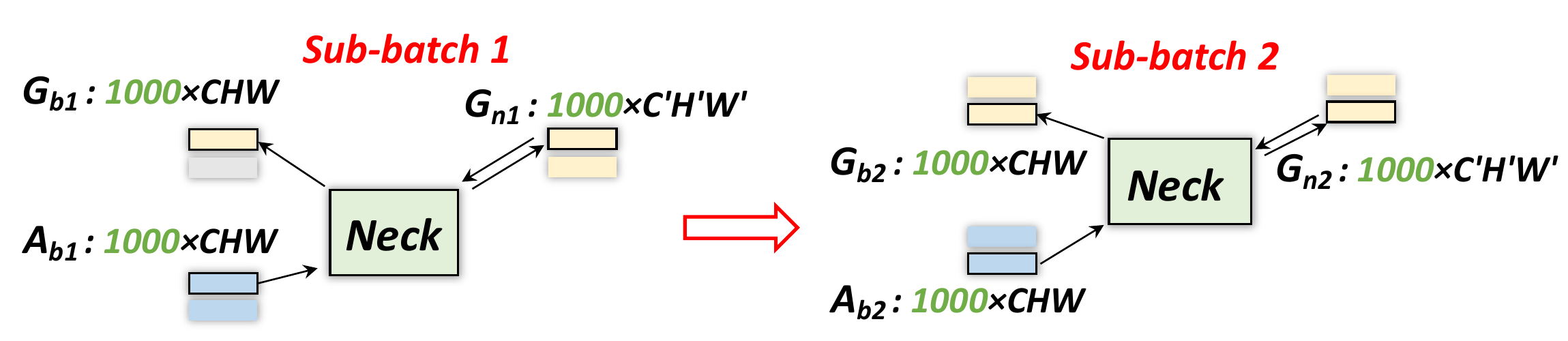}
\vspace{-2.5em}
\captionsetup{labelformat=empty}
\caption{(b) Split $A_b, G_n$ into sub-batches to update `Neck'. \\$G_b$ accumulated.}
\end{minipage}
\begin{minipage}{.19\textwidth}
\centering
\includegraphics[width=\linewidth]{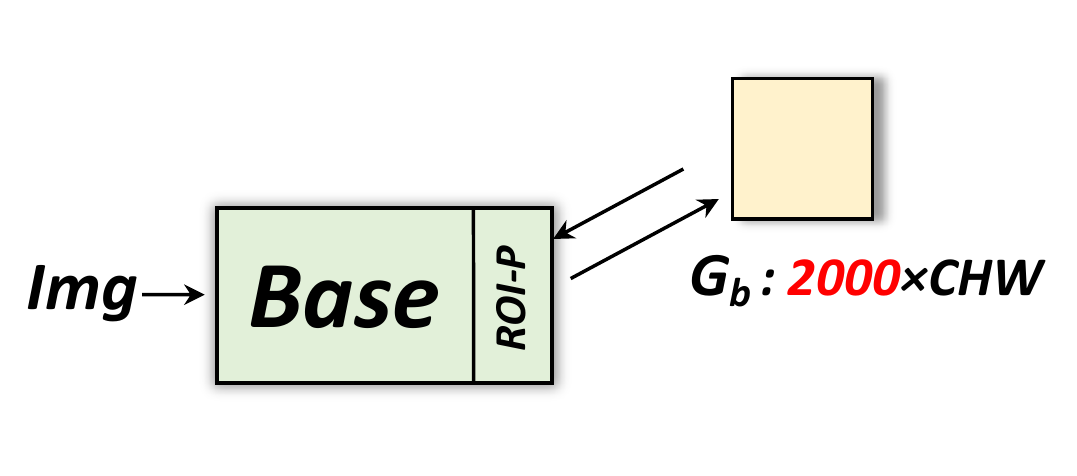}
\vspace{-2em}
\captionsetup{labelformat=empty}
\caption{(c) Use $G_b$ to update `Base' network. }
\end{minipage}
\vspace{-1em}
\caption{Seq-BBP:  {\color{blue}blue}, {\color{yellow}yellow}, and {\color{green}green} blobs represent activation, gradients, and the module that is being updated.}
\label{fig:step-bp}
\vspace{-1.5em}
\end{figure*}

Because of the intra-category variation, existing WSOD methods often mistakenly only detect the discriminative parts of an object rather than its full extent. A natural solution for this issue encourages the network to focus on the context which can be achieved by dropping the most discriminative parts. Hence, spatial dropout is an intuitive fit. 

Na\"ive spatial dropout has limition for detection since the discriminative parts of objects differ in location and size. A more structured DropBlock~\cite{dropblock} was proposed where spatial points on ROI feature maps are sampled randomly as blob centers, and the square regions around these centers of size $H\times H$ are then dropped across all channels on the ROI feature map. Finally, the feature values are re-scaled by a factor of the area of the whole ROI over the area of the un-dropped region so that no normalization has to be applied for inference when no regions are dropped. 

DropBlock is a non-parametric regularization technique. While it is able to improve model robustness and alleviate part domination, it basically treats regions equally. We consider dropping more frequently at discriminative parts in an adversarial manner. To this end, we develop the \textit{Concrete DropBlock}: a data-driven and parametric variant of DropBlock which is learned end-to-end to drop the most relevant regions as shown in Fig.~\ref{fig:dropblock}. Given an input image, the feature maps $\psi_w(r)\in\mathbb{R}^{H\times H}$ are computed for each region $r \in R$ using the layers up until ROI-Pooling. $H$ is the ROI-Pooling output dimension. We then feed $\psi_w(r)$ into a convolutional residual block to generate a probability map $p_\theta(r)\in\mathbb{R}^{H\times H}~\forall r \in R$ where $\theta$ subsumes the trainable parameters of this module. Each element of $p_\theta(r)$ is regarded as an independent Bernoulli variable, and this probability map is transformed via a spatial Gumbel-Softmax~\cite{Gumbel-Softmax,concrete} into a hard mask $M_\theta(r) \in \{0,1\}^{H\times H}~\forall r \in R$. This operation is a differentiable approximation of  sampling. To avoid trivial solutions (\eg, everything will be dropped or a certain area is dropped consistently), we apply a threshold $\tau$ such that $p_\theta(r) = \min (p_\theta(r), \tau)$. This  guarantees that the computed mask $M_\theta(r)$ is sparse. We follow DropBlock to finally generate the structured mask and normalize the features. During training, we jointly optimize the original network parameters $w$ and the residual block parameters $\theta$ with the following minmax objective:
\begin{equation}
\small{
w^*, \theta^* = \arg \min_w \max_\theta \sum_{I} \mathcal{L}_\text{img} (w, \theta) +  \mathcal{L}_\text{roi} (w, \theta). 
}
\end{equation}
By maximizing the original loss \wrt the Concrete DropBlock parameters, the Concrete DropBlock will learn to drop the most discriminative parts of the objects, as it is  the easiest way to increase the training loss. This forces the object detector  to also look at the context regions. We found this strategy to improve performance especially for  non-rigid object categories, which usually have a large intra-class difference.

\subsection{Sequential batch back-propagation}
\label{sec:step-bp}
In this section, we discuss how we propose to handle memory limitations particularly during training, which turn out to be a major bottleneck preventing previous WSOD methods from using state-of-the-art deep nets. We introduce our  memory-efficient \emph{sequential batch forward and backward computation}, tailored for WSOD models. 

Vanilla training via back-propagation~\cite{Rumelhart}  stores all intermediate activations during the forward pass, which are  reused when computing gradients of network parameters. This method is computationally efficient due to memoization, yet memory-demanding for the same reason. More efficient versions~\cite{Kokkinos17,ChenXZG16} have been proposed, where only a subset of the intermediate activations are saved during a forward pass at key layers. The whole model is cut into smaller sub-networks at these key layers. When computing gradients for a sub-network, a forward pass is first applied  to obtain the intermediate representations for this sub-network, starting from the stored activation at the input key layer of the sub-network. Combined with the gradients propagated from earlier sub-networks, the gradients of sub-network weights are computed and gradients are also propagated to outputs of earlier sub-networks.

This algorithm is designed for extremely deep networks where the memory cost is roughly evenly distributed along the layers. However, when these deep nets are adapted for detection, the activations (after ROI-Pooling) grow from $1\times CHW$ (image feature) to $N\times CHW$ (ROI-features) where $N$ is in the thousands for weakly supervised models. Without ground-truth boxes, all these proposals  need to be maintained for high recall and thus good performance (see the evidence in Appendix~\ref{app:prop}).

To address this training challenge, we propose a sequential computation in the `Neck' sub-module as depicted in Fig.~\ref{fig:step-bp}. During the forward pass, the input image is first passed through the `Base' and `Neck,' with only the activation $A_b$ after the `Base' stored. The output of the `Neck' then goes into the `Head' for its first forward and backward pass to update the weights of the `Head' and the gradients $G_n$  as shown in Fig.~\ref{fig:step-bp}~(a). To update the parameters of the `Neck,' we split the ROI-features into `sub-batches' and run back-propagation on each small sub-batch sequentially. Hence we avoid storing memory-consuming feature maps and their gradients within the `Neck.' An example of this sequential method is shown in Fig.~\ref{fig:step-bp}~(b), where we split 2000 proposals into two sub-batches of 1000 proposals each. The gradient $G_b$ is accumulated and used to  update the parameters of the `Base' network via regular back-propagation as illustrated in Fig.~\ref{fig:step-bp}~(c).  For testing, the same strategy can be applied if either the number of ROIs or the size of the `Neck' is too large.

% !TEX root = ./main.tex
\section{Experiments}
\label{sec:exp}

We assess our proposed method subsequently after detailing dataset, evaluation metrics and implementation.

\begin{table}[t]
\centering
\resizebox{\columnwidth}{!}{
%  \footnotesize{
\begin{tabular}{c| c c | c c }
\specialrule{.15em}{.05em}{.05em}
Methods  & Val-AP & Val-AP$_{50}$   & Test-AP & Test-AP$_{50}$  \\
\hline
Fast R-CNN   & 18.9 & 38.6  & 19.3 & 39.3  \\
Faster R-CNN  & 21.2 & 41.5  & 21.5 & 42.1 \\
\hline
WSDDN~\cite{Bilen16}   & - & -  & - & 11.5\\
WCCN~\cite{DibaSPPG17} & - & -  & - & 12.3 \\
PCL~\cite{tang2018pcl}  & 8.5 & 19.4 \\
C-MIDN~\cite{Gao_2019_ICCV}  & 9.6 & 21.4  & - & - \\
WSOD2~\cite{Zeng_2019_ICCV} & 10.8 & 22.7 & - & - \\
\hline
Diba~\etal\cite{diba-adv}+SSD   & - & -  & - & 13.6 \\
OICR~\cite{tang2017multiple}+Ens+FRCNN  & 7.7 & 17.4 & - & - \\
Ge~\etal\cite{Ge_2018_CVPR}+FRCNN  & 8.9 & 19.3 & - & -  \\
PCL~\cite{tang2018pcl}+Ens.+FRCNN  & 9.2 & 19.6 & - & - \\
\hline
Ours (single-model)  & \textbf{11.4} & \textbf{24.3} & \textbf{12.1} & \textbf{24.8}   \\
\specialrule{.15em}{.05em}{.05em}
\end{tabular}
}
\vspace{-1em}
\caption{Single model results (VGG16) on COCO.}
\label{table:coco}
\vspace{0.3em}
\centering
% \resizebox{\columnwidth}{!}{
 \footnotesize{
\begin{tabular}{c c c |  c c }
\specialrule{.15em}{.05em}{.05em}
Methods & Proposal & Backbone  & AP  & AP$_{50}$ \\
\hline
Faster R-CNN & RPN & R101-C4   & 27.2 & 48.4 \\
\hline
Ours & MCG & VGG16    & \textbf{11.4} & \textbf{24.3}  \\
Ours & MCG & R50-C4   & \textbf{12.6} & \textbf{26.1}  \\
Ours & MCG & R101-C4 & \textbf{13.0} & \textbf{26.3}  \\
\specialrule{.15em}{.05em}{.05em}
\end{tabular}
}
\vspace{-1em}
\caption{Single model results (ResNet) on COCO 2014 val.}
\label{table:voc-coco-res}
\vspace{-1em}
\end{table}

\vspace{-1em}
\paragraph{\textbf{Dataset and evaluation metrics.} }
We first conduct experiments on COCO~\cite{coco}, which is the most popular dataset used for supervised object detection but rarely studied in WSOD. We use the COCO 2014 train/val/test split and report standard COCO metrics including AP (averaged over IoU thresholds) and AP$_{50}$ (IoU threshold at 50\%).  

We then evaluate on both VOC 2007 and 2012~\cite{pascal}, which are commonly used to assess WSOD performance. Average Precision (AP) with IoU threshold at 50\% is used to evaluate the accuracy of object detection (Det.) on the testing data. We also evaluate correct localization accuracy (CorLoc.), which measures the percentage of training images of a class for which the most confident predicted box has at least 50\% IoU with at least one ground-truth box.

\vspace{-1em}
\paragraph{\textbf{Implementation details.} }
For a fair comparison, all  settings of the VGG16 model are kept identical to~\cite{tang2017multiple, tang2018pcl} except those mentioned below. We use 8 GPUs during training with one input image per device. SGD is used for optimization. The default $p$ and IoU in our proposed MIST technique (Alg.~\ref{alg:inst-or}) are set to $0.15$ and $0.2$. For the Concrete DropBlock $\tau = 0.3$. The ResNet models are  identical to~\cite{fastrcnn}. Please check Appendix~\ref{app:implement} for more  details.

% \vspace{-1em}
\subsection{Overall performance}

\noindent\textbf{VGG16-COCO.}
We compare to state-of-the-art WSOD methods on COCO in Tab.~\ref{table:coco}. Our single model without any post-processing outperforms all previous approaches (w/ bells and whistles) by a great margin. On the private Test-dev benchmark, we increase  AP$_{50}$ by 11.2 (+82.3\%). For the 2014 validation set, we increase  AP and AP$_{50}$ by 0.6 (+5.6\%) and 1.6 (+7.1\%). Complete results are provided in Appendix~\ref{app:coco}.  Note that compared to  supervised models  shown in the first two rows, the performance gap is still relatively big: ours is 56.9\% of Faster R-CNN on average. In addition, our model achieves 12.4 AP and 25.8 AP$_{50}$ on the COCO 2017 split as reported in Tab.~\ref{table:more_data}, which is more commonly adopted in supervised papers. 

\begin{figure*}[h]
\vspace{-1em}
\centering
\includegraphics[width=0.99\textwidth]{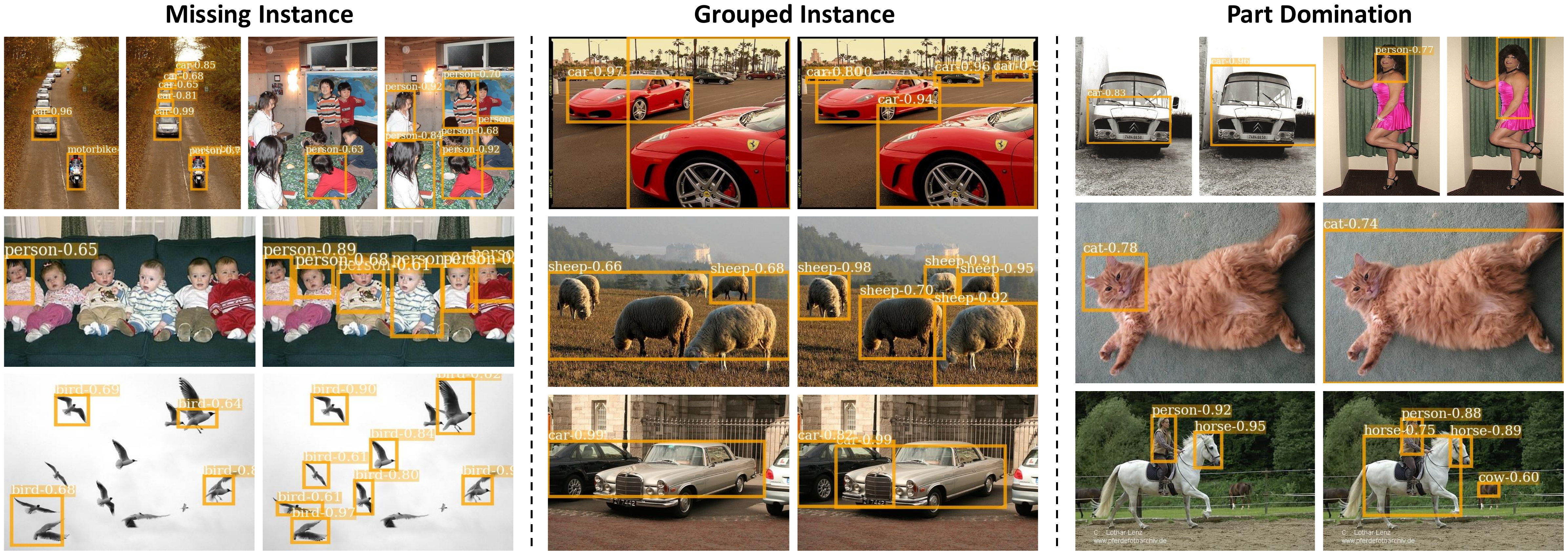}
\vspace{-0.93em}
\caption{Comparison of our models (right picture in pair) to our  baseline (left picture in pair).}
\label{fig:quali1}
\vspace{0.2em}
\includegraphics[width=0.99\textwidth]{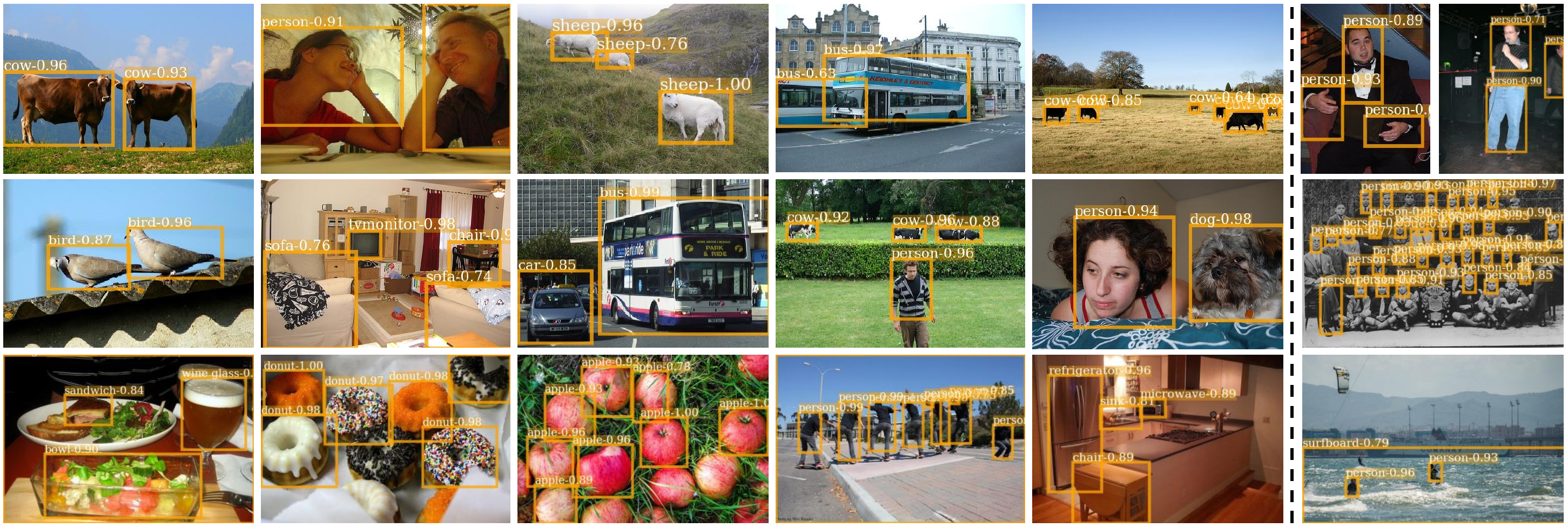}
\vspace{-0.93em}
\caption{More visualization (top: VOC 2007, middle: VOC 2012, bottom: COCO) and some failure cases (right column).}
\label{fig:quali2}
\vspace{-1.0em}
\end{figure*}

\noindent\textbf{ResNet-COCO.} ResNet models have never been trained and evaluated before for WSOD. Nonetheless, they are the most popular backbone networks for supervised methods. Part of the reason is the larger memory consumption of ResNet. Without the training techniques introduced in Sec.~\ref{sec:step-bp}, it's impossible to train  on a standard GPU using all proposals. In Tab.~\ref{table:voc-coco-res} we provide the first benchmark for the COCO dataset using ResNet-50 and ResNet-101. As expected we observe ResNet models to perform better than the VGG16 model. Moreover, we note that the difference between ResNet-50 and ResNet-101 is relatively small. 

\noindent\textbf{VGG16-VOC.}
To fairly compare with most previous WSOD works, we also evaluate our approach on the VOC datasets~\cite{pascal}. The comparison to  most recent works is reported in Tab.~\ref{table:voc-single}. All entries in this table are single model results. For object detection, our single-model results surpass all previous approaches on the publicly available 2007 test set (+1.3 AP$_{50}$) and on the private 2012 test set (+1.9 AP$_{50}$). In addition, our single model also performs better than all  previous methods with bells and whistles (\eg, `+FRCNN': supervised re-training, `+Ens.': model ensemble). Combining the 2007 and 2012  training set, our model  achieves 58.1\% (+2.1 AP$_{50}$) on the 2007 test set as reported in Tab.~\ref{table:more_data}. CorLoc results on the training set and  per-class results are provided in Appendix~\ref{app:voc}. Since VOC is  easier than COCO, the  performance gap to supervised methods is smaller: ours is 78.1\% of Faster R-CNN on average.

\begin{table}[t]
\centering
% \resizebox{\columnwidth}{!}{
 \footnotesize{
\begin{tabular}{c | c | c | c}
\specialrule{.15em}{.05em}{.05em}
Methods & Proposal & 07-AP$_{50}$ & 12-AP$_{50}$ \\
\hline
Fast R-CNN & SS           & 66.9   &  65.7  \\
Faster R-CNN  & RPN       & \textbf{69.9}  &  \textbf{67.0}  \\
\hline
WSDDN~\cite{Bilen16}      & EB  & 34.8 & - \\
OICR~\cite{tang2017multiple} & SS & 41.2 & 37.9 \\
PCL~\cite{tang2018pcl}    & SS  & 43.5  & 40.6 \\
SDCN~\cite{Li_2019_ICCV} & SS & 50.2 & 43.5 \\
Yang \etal~\cite{Yang_2019_ICCV} & SS  & 51.5  & 45.6 \\
C-MIL~\cite{c-mil}        & SS  & 50.5  & 46.7 \\
WSOD2~\cite{Zeng_2019_ICCV} & SS  & \textbf{53.6} & 47.2 \\
Pred Net~\cite{Arun_2019}   & SS  & 52.9  & 48.4 \\ 
C-MIDN~\cite{Gao_2019_ICCV} & SS & 52.6  & \textbf{50.2} \\ 
\hline
C-MIL~\cite{c-mil}+FRCNN &  SS & 53.1  & - \\
SDCN~\cite{Li_2019_ICCV}+FRCNN & SS & 53.7 & 46.7 \\
Pred Net~\cite{Arun_2019}+Ens.+FRCNN & SS  & 53.6  & 49.5 \\
Yang \etal~\cite{Yang_2019_ICCV}+Ens.+FRCNN & SS  & 54.5  & 49.5 \\
C-MIDN~\cite{Gao_2019_ICCV}+FRCNN  & SS  & 53.6  & 50.3 \\
\hline
Ours (single) & SS & \textbf{54.9} & \textbf{52.1\footnotemark}  \\
\specialrule{.15em}{.05em}{.05em}
\end{tabular}
}
\vspace{-1em}
\caption{Single model (VGG16) detection results on VOC.}
\label{table:voc-single}
\vspace{-1em}
\end{table}
\footnotetext{\scriptsize{\url{http://host.robots.ox.ac.uk:8080/anonymous/DCJ5GA.html}}}

\noindent\textbf{Additional training data.} 
The biggest advantage of WSOD methods is the availability of more data. Therefore, we are interested in studying whether more training data  improves results. We train our model on the VOC 2007 trainval (5011 images), 2012 trainval (11540 images), and the combination of both (16555 images) separately, and evaluate on the VOC 2007 test set. As shown in Tab.~\ref{table:more_data} (top), the performance increase consistently with the amount of training data. We verify this on COCO where 2014-train (82783 images) and 2017-train (128287 images) are used for training, and 2017-val (a.k.a.\ minival) for testing. Similar results are observed as shown in Tab.~\ref{table:more_data} (bottom). 

\begin{table}[t]
\centering
 \footnotesize{
\begin{tabular}{c | c c c c}
\specialrule{.15em}{.05em}{.05em}
Data-Split & 07-Trainval & 12-Trainval & 07-Test  \\
Metrics       & CorLoc      & CorLoc  & Det \\
\hline
Ours-07       & 68.8 & -    & 54.9 \\
Ours-12       & -    & 70.9 & 56.3 \\
WSOD2(07+12)~\cite{Zeng_2019_ICCV}  & 71.4  & 72.2 & 56.0 \\
Ours-(07+12)  & \textbf{71.8} & \textbf{72.9} & \textbf{58.1} \\
\specialrule{.15em}{.05em}{.05em}
Metrics     & 17-Val-AP  & 17-Val-AP$_{50}$ & 17-Val-AP$_{75}$ \\
\hline
Ours-Train2014  & 11.4 & 24.3 &  9.4 \\
Ours-Train2017  & \textbf{12.4} & \textbf{25.8} & \textbf{10.5} \\
\specialrule{.15em}{.05em}{.05em}
\end{tabular}
}
\vspace{-1em}
\caption{Does more data help?}
\label{table:more_data}
\vspace{-2em}
\end{table}

\subsection{Qualitative results}
Qualitatively, we compare our full model with Tang \etal~\cite{tang2017multiple}. In Fig.~\ref{fig:quali1} we show a set of two pictures side by side, with baselines on the left and our results on the right. Our model is able to address instance ambiguity by: (1) detecting previously ignored instances (Fig.~\ref{fig:quali1} left); (2) predicting tight and precise boxes for multiple instances instead of a big one (Fig.~\ref{fig:quali1} center). Part domination is also alleviated since our model focuses on the full extent of objects (Fig.~\ref{fig:quali1} right). Even though our model can greatly increase the score of larger boxes (see the horse example), the predictions may still be dominated by parts in some difficult cases.

More qualitative results are shown in Fig.~\ref{fig:quali2} for all three datasets we used, as well in Appendix~\ref{app:fig}. Our model is able to detect multiple instances of the same category (cow, sheep, bird, apple, person) and various objects of different classes (food, furniture, animal) in relatively complicated scenes. The COCO dataset is much harder than VOC as the number of objects and classes is bigger. Our model still tells apart objects decently well (Fig.~\ref{fig:quali2} bottom row).  We also show some failure cases (Fig.~\ref{fig:quali2} right column) of our model which can be roughly categorized into three types: (1) relevant parts are predicted as instances of objects (hands and legs, bike wheels); (2) in extreme examples, part domination remains (model converges to a face detector); (3) object co-occurrence  confuses the detector when it predicts the sea as a surfboard or the baseball court as a bat. 

\vspace{-1em}
\subsection{Analysis}
\label{exp:ablation}
\noindent\textbf{How much does each module help?} 
We study the effectiveness of each module in Tab.~\ref{table:module-abla}. We first reproduce the method of Tang \etal~\cite{tang2017multiple}, achieving similar results (first two rows). Applying the developed MIST module improves the results significantly. This aligns with our observation that instance ambiguity is the biggest bottleneck for WSOD. Our conceptually simple solution also outperforms an improved version~\cite{tang2018pcl} (PCL), which is based on a computationally expensive and carefully-tuned clustering. 

The devised Concrete DropBlock further improves the performance when using MIST as the basis. This module  surpasses several variants  including:  (1) (Img Spa.-Dropout): spatial dropout applied on the image-level features; (2) (ROI-Spa.-Dropout): spatial dropout applied on each ROI where each feature point is treated independently. This setting is similar to~\cite{singh-iccv2017, a-fast-rcnn}; (3) (DropBlock): the best-performing DropBlock setting reported in~\cite{dropblock}.

\begin{table}[t]
\centering
\resizebox{\columnwidth}{!}{
%  \footnotesize{
\begin{tabular}{c | c  c |  c c }
\specialrule{.15em}{.05em}{.05em}
Data-Split & 07 trainval    & 07 test & 12 trainval    & 12 test  \\
Metrics & CorLoc &  Det.  & CorLoc &  Det.   \\
\hline
Baseline~\cite{tang2017multiple}*  & 60.8 & 42.5  & - & - \\
\hline
$+$ PCL~\cite{tang2018pcl} & 62.7 & 43.5 & 63.2 & 40.6  \\
$+$ MIST w/o Reg. & 62.9 & 48.3 & 65.1 & - \\ 
$+$ MIST  & \textbf{64.9}  &  \textbf{51.4}  & \textbf{66.7}  & -  \\
\hline
$+$ Img Spa.-Dropout & 64.3  & 51.1 & 65.9 &  -\\
$+$ ROI Spa.-Dropout & 66.8 & 52.4 & 67.3 &  -\\
$+$ DropBlock~\cite{dropblock}    & 67.1 & 52.9 &  68.4  & - \\
$+$ Concrete DropBlock &  \textbf{68.8}  &   \textbf{54.9} & \textbf{70.9}  & \textbf{52.1} \\
\specialrule{.15em}{.05em}{.05em}
\end{tabular}
}
\vspace{-1em}
\caption{Ablation study. (*: our implementation)}
\label{table:module-abla}
\vspace{0.5em}
\centering
\resizebox{\columnwidth}{!}{
 \footnotesize{
\begin{tabular}{c | c  c c c c c }
\specialrule{.15em}{.05em}{.05em}
Metrics &  $AR^{1}$  &  $AR^{10}$ &  $AR^{100}$ & $AR^{\text{s}}$  &  $AR^{\text{m}}$  &  $AR^{\text{l}}$ \\
\hline
w/o MIST & 18.6 & 30.6 & 32.5 & 8.8 & 25.8 & 38.9 \\
w/  MIST &  \textbf{20.5} & \textbf{37.8} & \textbf{43.9} &  \textbf{15.0} &  \textbf{34.8} &  \textbf{51.7} \\
\specialrule{.15em}{.05em}{.05em}
\end{tabular}
}}
\vspace{-1em}
\caption{Average Recall (AR) (\%) comparison.}
\label{table:ar}
\vspace{-1em}
\end{table}

\vspace{-1em}
\paragraph{\textbf{Has Instance Ambiguity been addressed?}}
To validate that instance ambiguity is alleviated, we report Average Recall (AR) over multiple IoU values ($.50:.05:.95$), given 1, 10, 100 detections per image ($AR^{1}$, $AR^{10}$, $AR^{100}$) and for small, medium, annd large objects ($AR^{\text{s}}$, $AR^{\text{m}}$, $AR^{\text{l}}$) on VOC 2007. We compare the model with and without MIST in Tab.~\ref{table:ar} where our method increases all recall metrics. 

\begin{figure}[t]
\centering
\begin{subfigure}{0.49\linewidth}
\centering\includegraphics[width=1\linewidth]{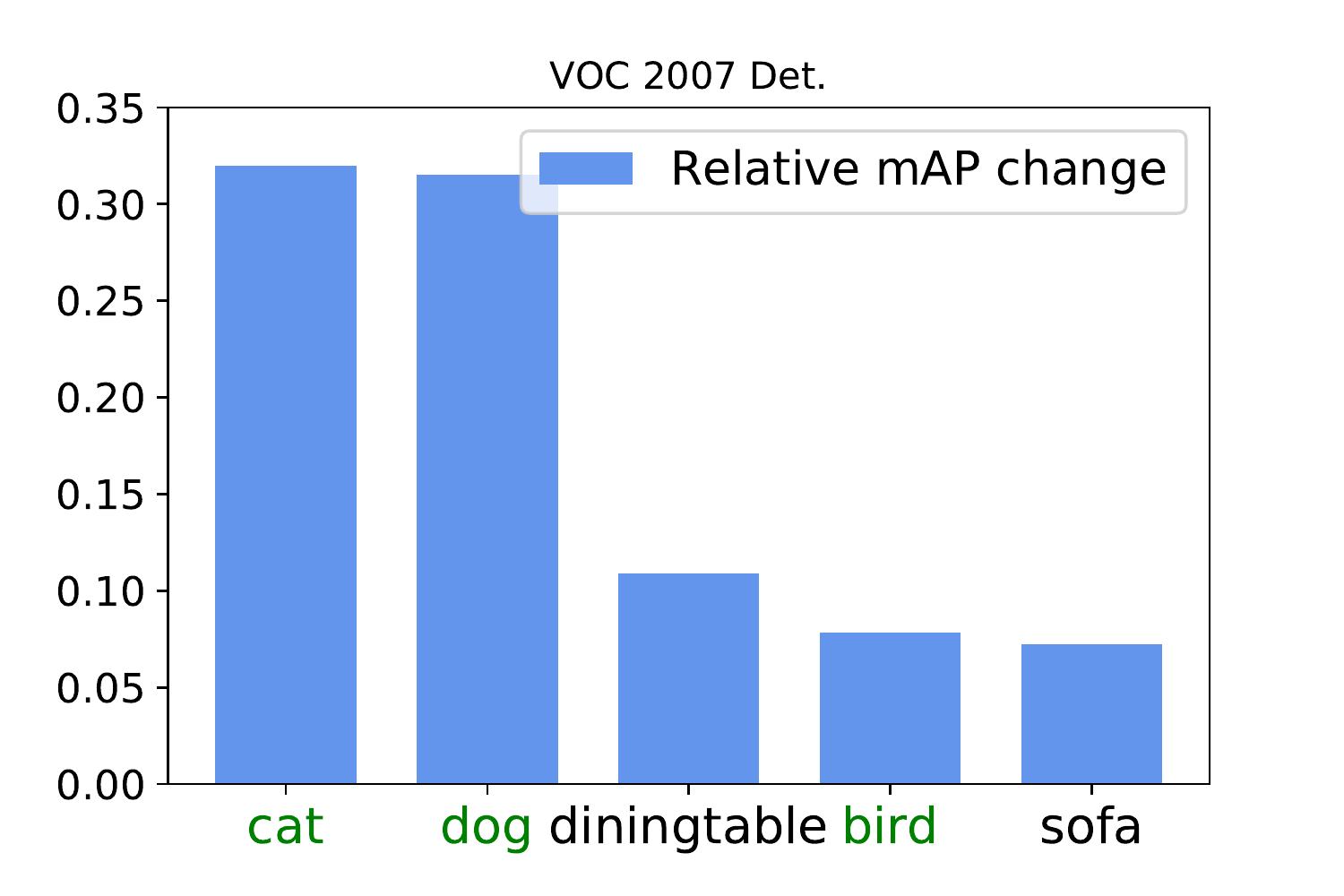}
\end{subfigure}%
\begin{subfigure}{0.49\linewidth}
\centering\includegraphics[width=1\linewidth]{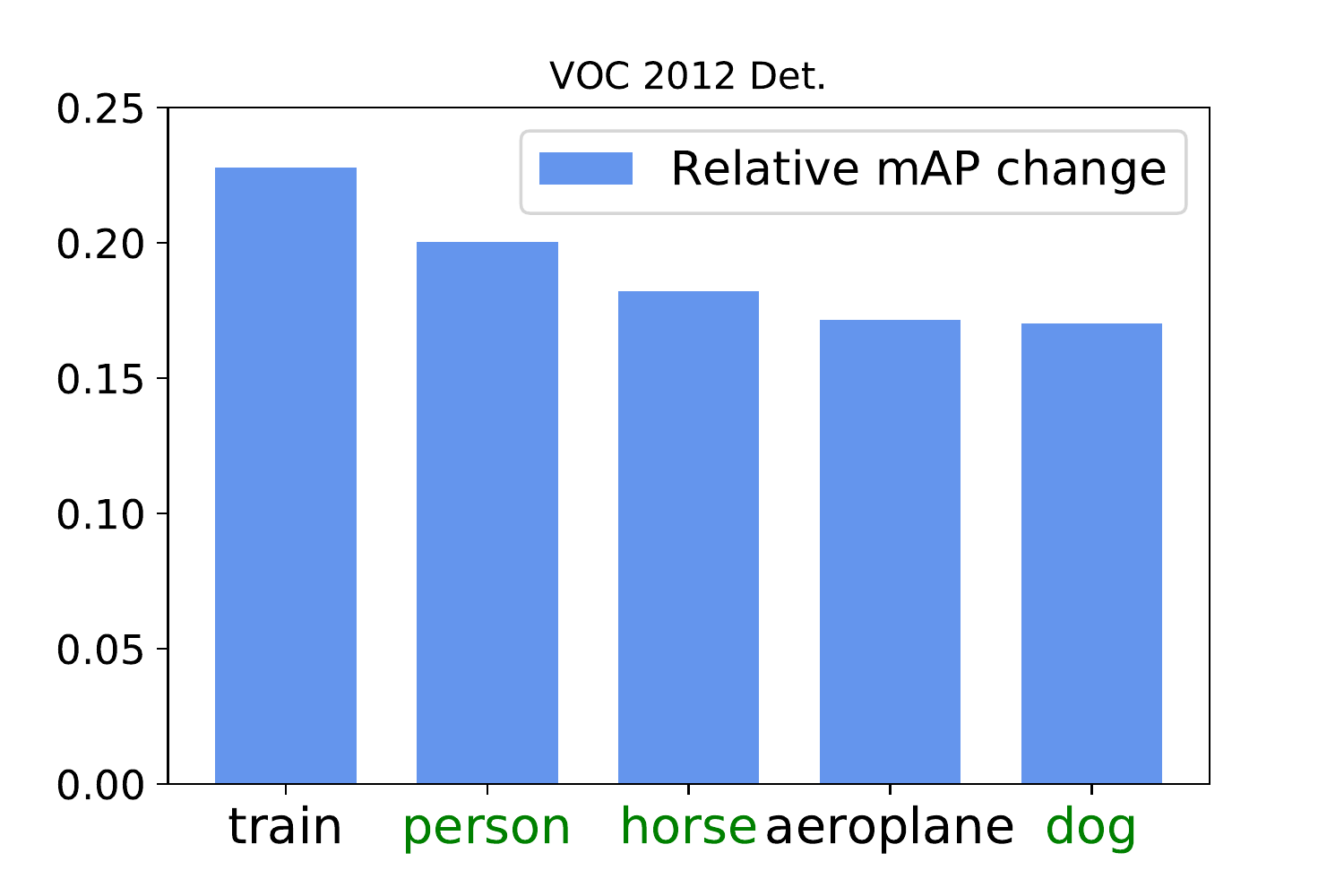}
\end{subfigure}
\vspace{-1em}
\caption{Top-5 classes with biggest performance boost when using Concrete DropBlock. Animal classes are emphasized using green color.}
\label{fig:per-cls}
\vspace{0.5em}
\centering
\includegraphics[width=0.42\textwidth]{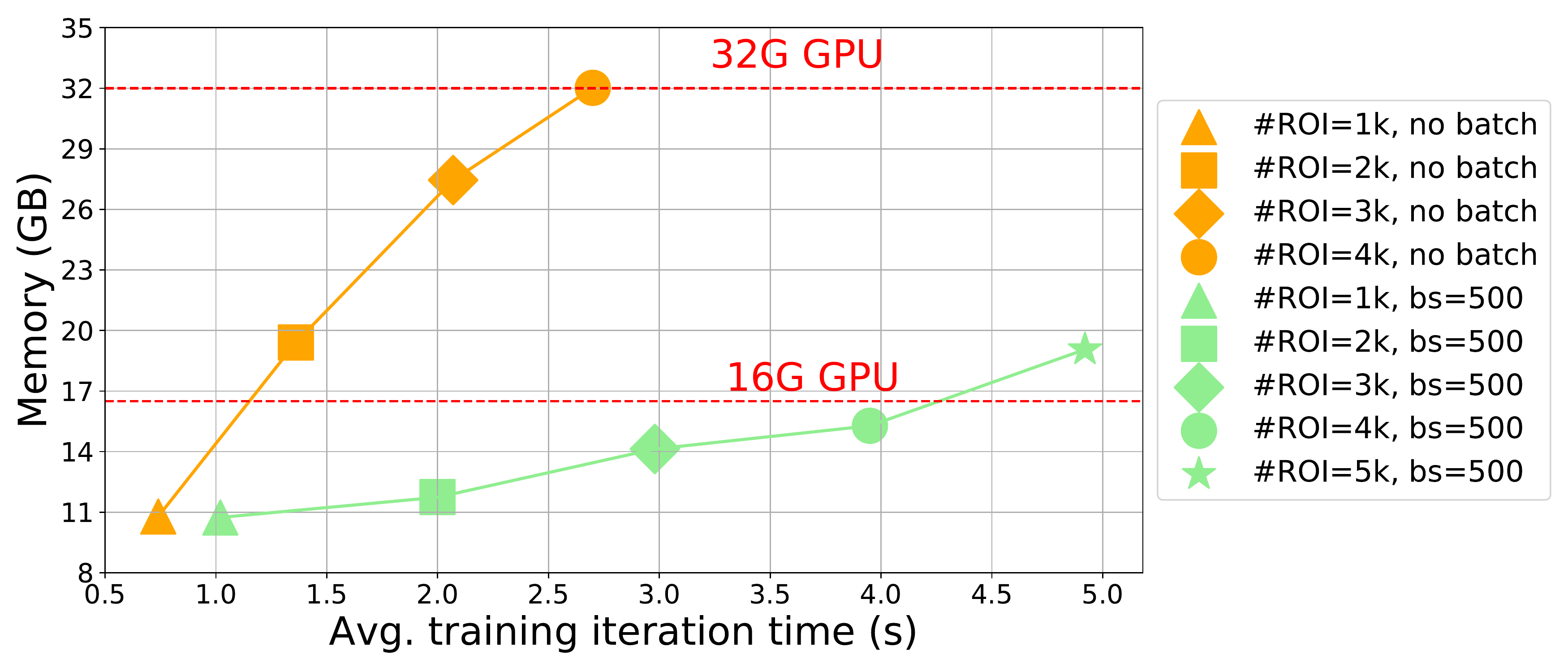}
\vspace{-1em}
\caption{ResNet-101 model memory consumption using different methods and different number of proposals.} 
\label{fig:mem-exp}
\vspace{-1em}
\end{figure}

\vspace{-1em}
\paragraph{\textbf{Has Part Domination been addressed?}}
In Fig.~\ref{fig:per-cls}, we show the 5 categories with the biggest relative performance improvements on the VOC 2007 and VOC 2012 dataset after applying the Concrete DropBlock. The performance of animal classes including `person' increases most, which matches our intuition mentioned in Sec.~\ref{sec:intro}: the part domination issue is most prominent for articulated classes with rigid and discriminative parts. Across both datasets, three out of the five top classes are mammals.

\vspace{-1em}
\paragraph{\textbf{Space-time analysis of sequential batch BP?} }
We also study the effect of our sequential batch back-propagation. We fix the input image to be of size $600 \times 600$, and run two methods (vanilla back-propagation and ours with sub-batch size 500 %\A{what's `sub-batch' size? not defined?} 
using ResNet-101 for comparison. We change the number of proposals from 1k to 5k in 1k increments, and report average training iteration time and memory consumption in Fig.~\ref{fig:mem-exp}. We observe: (1)  vanilla back-propagation cannot even afford 2k proposals (average number of ROIs widely used in~\cite{fastrcnn, Bilen16, tang2017multiple}) on a standard 16GB GPU, but ours can easily handle up to 4k boxes; (2) the training process is not greatly slowed down, ours takes $\sim$1-2$\times$ more time than the vanilla version. In practice,  input resolution and total number of proposals can be  bigger. 

\begin{table}[t]
\centering
% \resizebox{\columnwidth}{!}{
 \footnotesize{
\begin{tabular}{c | c  c | c c }
\specialrule{.15em}{.05em}{.05em}
Methods & Backbone   & Det. (AP) & Backbone & Det. (AP) \\
\hline
Supervised  & VGG16  & 61.7~\cite{xiao-eccv2018} & R-101 & 80.5~\cite{xiao-eccv2018} \\
\hline
\cite{Bilen16} & VGG16   & 24.2 & R-101   &  21.9 \\
\cite{tang2017multiple}  & VGG16  & 34.8  & R-101 & 40.5 \\
\hline
Ours (MIST only) & VGG16  & 35.7 & R-101  & 44.0\\
Ours & VGG16   & \textbf{36.6} & R-101  &  \textbf{45.7} \\
\hline
Ours+flow & VGG16 &  \textbf{38.3} & R-101 & \textbf{46.9}  \\
\specialrule{.15em}{.05em}{.05em}
\end{tabular}}
\vspace{-1em}
\caption{Video Object Detection Results.}
\label{table:vid}
\vspace{-1em}
\end{table}

\begin{figure}[t]
\centering
\begin{subfigure}{0.49\linewidth}
\centering\includegraphics[width=1\linewidth]{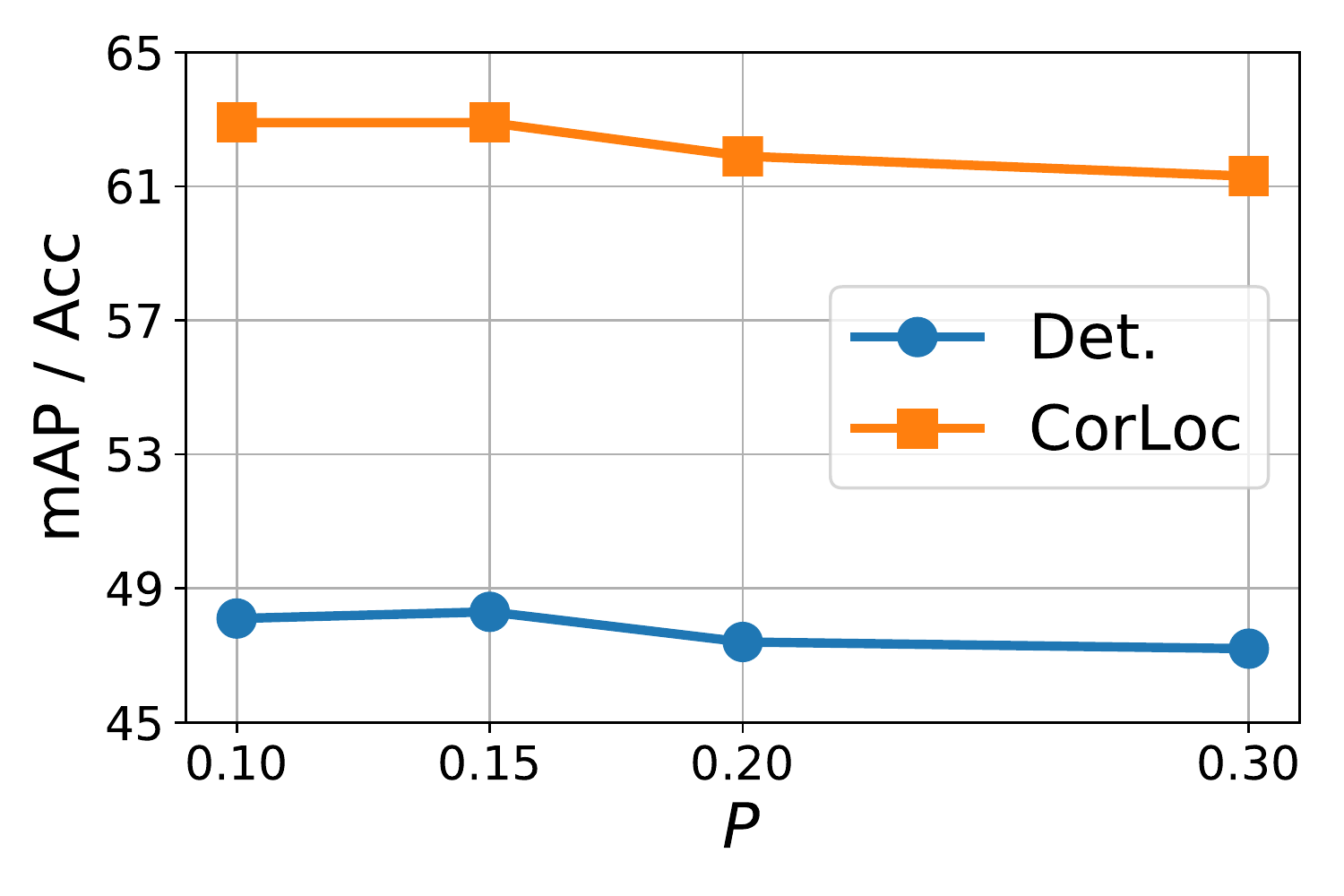}
\end{subfigure}%
\begin{subfigure}{0.49\linewidth}
\centering\includegraphics[width=1\linewidth]{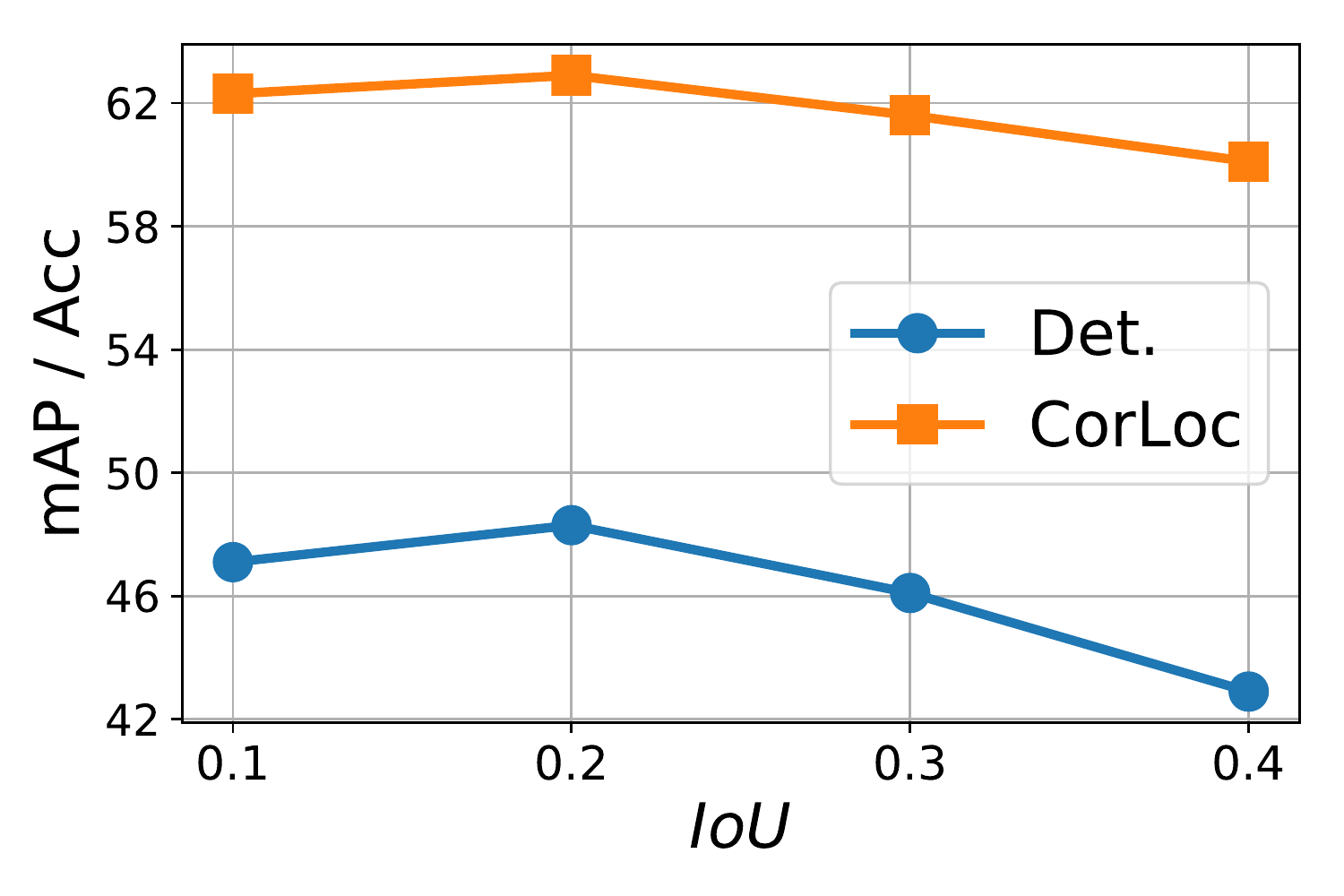}
\end{subfigure}
\vspace{-1em}
\caption{VOC 2007 results for different $p$ and IoU.}
\label{fig:oicr-abla}
\vspace{-1em}
\end{figure}

\vspace{-1em}
\paragraph{\textbf{Robustness of MIST?} }
To assess robustness we test a baseline model plus this algorithm only using different top-percentage $p$ and rejection IoU on the VOC 2007 dataset. Results are shown in Fig.~\ref{fig:oicr-abla}. The best result is achieved with $p=0.15$ and $IoU=0.2$, which we use for all the other models and datasets. Importantly, we note that, overall, the sensitivity of the final results  on the value of $p$ is small and only slightly larger for IoU. 

\subsection{Extension: video object detection}
\label{sec:video}

We finally generalize our models to video-WSOD, which hasn't been explored in the literature. Following supervised methods, we experiment on the most popular dataset: ImageNet VID~\cite{imagenet}. Frame-level category labels are available during training. Uniformly sampled key-frames are used for training following~\cite{zhu17fgfa} and evaluation settings are also kept identical. Results are reported in Tab.~\ref{table:vid}. The performance improvement of the proposed MIST and Concrete DropBlock  generalize to  videos. The memory-efficient sequential batch back-propagation permits to leverage  short-term motion patterns (\ie,  we use optical-flow following~\cite{zhu17fgfa}) to further increase the performance. This suggests that videos are a useful domain where we can obtain more data to improve WSOD. Full details are provided in Appendix~\ref{app:vid}.

\vspace{-0.5em}
\section{Conclusion}
In this paper, we address three major issues of WSOD. For each we have proposed a solution and demonstrated its effectiveness through extensive experiments. We achieve state-of-the-art results on popular datasets (COCO, VOC 07 and 12) and are the first to benchmark ResNet backbones and weakly supervised video object detection.

\vspace{0.3em}
\noindent\textbf{Acknowledgement:} ZR is supported by Yunni \& Maxine Pao Memorial Fellowship. This work is supported in part by NSF under Grant No.\ 1718221 and  No.\ 1751206.

{\small
\bibliographystyle{ieee_fullname}
\bibliography{egbib}

\begin{thebibliography}{10}\itemsep=-1pt

\bibitem{mcg}
P. Arbel\'{a}ez, J. Pont-Tuset, J. Barron, F. Marques, and J. Malik.
\newblock Multiscale combinatorial grouping.
\newblock In {\em Proc. CVPR}, 2014.

\bibitem{Arun_2019}
Aditya Arun, C.~V. Jawahar, and M.~Pawan Kumar.
\newblock Dissimilarity coefficient based weakly supervised object detection.
\newblock In {\em Proc. CVPR}, 2019.

\bibitem{Bilen14b}
H. Bilen, M. Pedersoli, and T. Tuytelaars.
\newblock Weakly supervised object detection with posterior regularization.
\newblock In {\em Proc. BMVC}, 2014.

\bibitem{BilenPT15}
Hakan Bilen, Marco Pedersoli, and Tinne Tuytelaars.
\newblock Weakly supervised object detection with convex clustering.
\newblock In {\em Proc. CVPR}, 2015.

\bibitem{Bilen16}
H. Bilen and A. Vedaldi.
\newblock Weakly supervised deep detection networks.
\newblock In {\em Proc. CVPR}, 2016.

\bibitem{ChenXZG16}
Tianqi Chen, Bing Xu, Chiyuan Zhang, and Carlos Guestrin.
\newblock Training deep nets with sublinear memory cost.
\newblock {\em CoRR}, abs/1604.06174, 2016.

\bibitem{CinbisVS14}
Ramazan~Gokberk Cinbis, Jakob~J. Verbeek, and Cordelia Schmid.
\newblock Multi-fold {MIL} training for weakly supervised object localization.
\newblock In {\em Proc. CVPR}, 2014.

\bibitem{imagenet}
J. Deng, W. Dong, R. Socher, L.-J. Li, K. Li, and L. Fei-Fei.
\newblock {ImageNet: A Large-Scale Hierarchical Image Database}.
\newblock In {\em Proc. CVPR}, 2009.

\bibitem{DibaSPPG17}
Ali Diba, Vivek Sharma, Ali~Mohammad Pazandeh, Hamed Pirsiavash, and Luc~Van
  Gool.
\newblock Weakly supervised cascaded convolutional networks.
\newblock In {\em Proc. CVPR}, 2017.

\bibitem{diba-adv}
Ali Diba, Vivek Sharma, Rainer Stiefelhagen, and Luc~Van Gool.
\newblock Object discovery by generative adversarial {\&} ranking networks.
\newblock 2017.

\bibitem{pascal}
M. Everingham, L. Van~Gool, C.~K.~I. Williams, J. Winn, and A. Zisserman.
\newblock The pascal visual object classes (voc) challenge.
\newblock In {\em Proc. IJCV}, 2010.

\bibitem{Gao_2019_ICCV}
Yan Gao, Boxiao Liu, Nan Guo, Xiaochun Ye, Fang Wan, Haihang You, and Dongrui
  Fan.
\newblock C-midn: Coupled multiple instance detection network with segmentation
  guidance for weakly supervised object detection.
\newblock In {\em Proc. ICCV}, 2019.

\bibitem{Ge_2018_CVPR}
Weifeng Ge, Sibei Yang, and Yizhou Yu.
\newblock Multi-evidence filtering and fusion for multi-label classification,
  object detection and semantic segmentation based on weakly supervised
  learning.
\newblock In {\em Proc. CVPR}, 2018.

\bibitem{dropblock}
Golnaz Ghiasi, Tsung{-}Yi Lin, and Quoc~V. Le.
\newblock Dropblock: {A} regularization method for convolutional networks.
\newblock In {\em Proc. NIPS}, 2018.

\bibitem{Detectron2018}
Ross Girshick, Ilija Radosavovic, Georgia Gkioxari, Piotr Doll\'{a}r, and
  Kaiming He.
\newblock Detectron.
\newblock \url{https://github.com/facebookresearch/detectron}, 2018.

\bibitem{fastrcnn}
Ross~B. Girshick.
\newblock Fast {R-CNN}.
\newblock In {\em Proc. ICCV}, 2015.

\bibitem{rcnn}
Ross~B. Girshick, Jeff Donahue, Trevor Darrell, and Jitendra Malik.
\newblock Rich feature hierarchies for accurate object detection and semantic
  segmentation.
\newblock In {\em Proc. CVPR}, 2014.

\bibitem{he2017maskrcnn}
Kaiming He, Georgia Gkioxari, Piotr Doll\'{a}r, and Ross Girshick.
\newblock Mask r-cnn.
\newblock In {\em Proc. ICCV}, 2017.

\bibitem{resnet}
Kaiming He, Xiangyu Zhang, Shaoqing Ren, and Jian Sun.
\newblock Deep residual learning for image recognition.
\newblock In {\em Proc. CVPR}, 2016.

\bibitem{flownet2}
E. Ilg, N. Mayer, T. Saikia, M. Keuper, A. Dosovitskiy, and T. Brox.
\newblock Flownet 2.0: Evolution of optical flow estimation with deep networks.
\newblock In {\em Proc. CVPR}, 2017.

\bibitem{Gumbel-Softmax}
Eric Jang, Shixiang Gu, and Ben Poole.
\newblock Categorical reparameterization with gumbel-softmax.
\newblock In {\em Proc. ICLR}, 2017.

\bibitem{JieWJFL17}
Zequn Jie, Yunchao Wei, Xiaojie Jin, Jiashi Feng, and Wei Liu.
\newblock Deep self-taught learning for weakly supervised object localization.
\newblock In {\em Proc. CVPR}, 2017.

\bibitem{KantorovOCL16}
Vadim Kantorov, Maxime Oquab, Minsu Cho, and Ivan Laptev.
\newblock Contextlocnet: Context-aware deep network models for weakly
  supervised localization.
\newblock In {\em Proc. ECCV}, 2016.

\bibitem{varitional-db}
Durk~P Kingma, Tim Salimans, and Max Welling.
\newblock Variational dropout and the local reparameterization trick.
\newblock In {\em Proc. NIPS}. 2015.

\bibitem{Kokkinos17}
Iasonas Kokkinos.
\newblock Ubernet: Training a universal convolutional neural network for low-,
  mid-, and high-level vision using diverse datasets and limited memory.
\newblock In {\em Proc. CVPR}, 2017.

\bibitem{Laine2017}
Samuli Laine and Timo Aila.
\newblock Temporal ensembling for semi-supervised learning.
\newblock In {\em Proc. ICLR}, 2017.

\bibitem{cornernet}
Hei Law and Jia Deng.
\newblock Cornernet: Detecting objects as paired keypoints.
\newblock In {\em Proc. ECCV}, 2018.

\bibitem{LiHLW016}
Dong Li, Jia{-}Bin Huang, Yali Li, Shengjin Wang, and Ming{-}Hsuan Yang.
\newblock Weakly supervised object localization with progressive domain
  adaptation.
\newblock In {\em Proc. CVPR}, 2016.

\bibitem{Li_2019_ICCV}
Xiaoyan Li, Meina Kan, Shiguang Shan, and Xilin Chen.
\newblock Weakly supervised object detection with segmentation collaboration.
\newblock In {\em Proc. ICCV}, 2019.

\bibitem{coco}
Tsung-Yi Lin, Michael Maire, Serge Belongie, James Hays, Pietro Perona, Deva
  Ramanan, Piotr Dollár, and C.~Lawrence Zitnick.
\newblock Microsoft coco: Common objects in context.
\newblock In {\em Proc. ECCV}, 2014.

\bibitem{ssd}
Wei Liu, Dragomir Anguelov, Dumitru Erhan, Christian Szegedy, Scott Reed,
  Cheng-Yang Fu, and Alexander~C. Berg.
\newblock Ssd: Single shot multibox detector.
\newblock In {\em ECCV}, 2016.

\bibitem{concrete}
Chris~J. Maddison, Andriy Mnih, and Yee~Whye Teh.
\newblock The concrete distribution: {A} continuous relaxation of discrete
  random variables.
\newblock In {\em Proc. ICLR}, 2017.

\bibitem{Peng_2018_CVPR}
Chao Peng, Tete Xiao, Zeming Li, Yuning Jiang, Xiangyu Zhang, Kai Jia, Gang Yu,
  and Jian Sun.
\newblock Megdet: A large mini-batch object detector.
\newblock In {\em Proc. CVPR}, 2018.

\bibitem{PleissCHLMW17}
Geoff Pleiss, Danlu Chen, Gao Huang, Tongcheng Li, Laurens van~der Maaten, and
  Kilian~Q. Weinberger.
\newblock Memory-efficient implementation of densenets.
\newblock {\em CoRR}, abs/1707.06990, 2017.

\bibitem{yolo}
Joseph Redmon, Santosh~Kumar Divvala, Ross~B. Girshick, and Ali Farhadi.
\newblock You only look once: Unified, real-time object detection.
\newblock In {\em Proc. CVPR}, 2016.

\bibitem{ren16faster}
Shaoqing Ren, Kaiming He, Ross Girshick, and Jian Sun.
\newblock Faster r-cnn: Towards real-time object detection with region proposal
  networks.
\newblock {\em TPAMI}, 2016.

\bibitem{Rumelhart}
D.~E. Rumelhart, G.~E. Hinton, and R.~J. Williams.
\newblock Parallel distributed processing: Explorations in the microstructure
  of cognition.
\newblock chapter Learning Internal Representations by Error Propagation. MIT
  Press, 1986.

\bibitem{Shen_2019_CVPR}
Yunhang Shen, Rongrong Ji, Yan Wang, Yongjian Wu, and Liujuan Cao.
\newblock Cyclic guidance for weakly supervised joint detection and
  segmentation.
\newblock In {\em Proc. CVPR}, 2019.

\bibitem{vgg}
Karen Simonyan and Andrew Zisserman.
\newblock Very deep convolutional networks for large-scale image recognition.
\newblock In {\em Proc. ICLR}, 2015.

\bibitem{singh-iccv2017}
Krishna~Kumar Singh and Yong~Jae Lee.
\newblock Hide-and-seek: Forcing a network to be meticulous for
  weakly-supervised object and action localization.
\newblock In {\em Proc. ICCV}, 2017.

\bibitem{singh-cvpr2019}
Krishna~Kumar Singh and Yong~Jae Lee.
\newblock You reap what you sow: Using videos to generate high precision object
  proposals for weakly-supervised object detection.
\newblock In {\em Proc. CVPR}, 2019.

\bibitem{krishna-cvpr2016}
Krishna~Kumar Singh, Fanyi Xiao, and Yong~Jae Lee.
\newblock Track and transfer: Watching videos to simulate strong human
  supervision for weakly-supervised object detection.
\newblock In {\em Proc. CVPR}, 2016.

\bibitem{siva_iccv13}
P. Siva and T. Xiang.
\newblock Weakly supervised object detector learning with model drift
  detection.
\newblock In {\em Proc. ICCV}, 2013.

\bibitem{song14slsvm}
Hyun~Oh Song, Yong~Jae Lee, Stefanie Jegelka, and Trevor Darrell.
\newblock Weakly-supervised discovery of visual pattern configurations.
\newblock In {\em Proc. NIPS}, 2014.

\bibitem{tang2018pcl}
Peng Tang, Xinggang Wang, Song Bai, Wei Shen, Xiang Bai, Wenyu Liu, and Alan
  Yuille.
\newblock {PCL}: Proposal cluster learning for weakly supervised object
  detection.
\newblock {\em TPAMI}, 2018.

\bibitem{tang2017multiple}
Peng Tang, Xinggang Wang, Xiang Bai, and Wenyu Liu.
\newblock Multiple instance detection network with online instance classifier
  refinement.
\newblock In {\em Proc. CVPR}, 2017.

\bibitem{TangWWYLHY18}
Peng Tang, Xinggang Wang, Angtian Wang, Yongluan Yan, Wenyu Liu, Junzhou Huang,
  and Alan~L. Yuille.
\newblock Weakly supervised region proposal network and object detection.
\newblock In {\em Proc. ECCV}, 2018.

\bibitem{TehRW16}
Eu~Wern Teh, Mrigank Rochan, and Yang Wang.
\newblock Attention networks for weakly supervised object localization.
\newblock In {\em Proc. BMVC}, 2016.

\bibitem{TompsonGJLB15}
Jonathan Tompson, Ross Goroshin, Arjun Jain, Yann LeCun, and Christoph Bregler.
\newblock Efficient object localization using convolutional networks.
\newblock In {\em Proc. CVPR}, 2015.

\bibitem{ss}
J.R.R. Uijlings, K.E.A. van~de Sande, T. Gevers, and A.W.M.Smeulders.
\newblock Selective search for object recognition.
\newblock {\em IJCV}, 2013.

\bibitem{c-mil}
Fang Wan, Chang Liu, Wei Ke, Xiangyang Ji, Jianbin Jiao, and Qixiang Ye.
\newblock {C-MIL:} continuation multiple instance learning for weakly
  supervised object detection.
\newblock In {\em Proc. CVPR}, 2019.

\bibitem{Wan_2018_CVPR}
Fang Wan, Pengxu Wei, Jianbin Jiao, Zhenjun Han, and Qixiang Ye.
\newblock Min-entropy latent model for weakly supervised object detection.
\newblock In {\em Proc. CVPR}, 2018.

\bibitem{WangRHT14}
Chong Wang, Weiqiang Ren, Kaiqi Huang, and Tieniu Tan.
\newblock Weakly supervised object localization with latent category learning.
\newblock In {\em Proc. ECCV}, 2014.

\bibitem{a-fast-rcnn}
Xiaolong Wang, Abhinav Shrivastava, and Abhinav Gupta.
\newblock A-fast-rcnn: Hard positive generation via adversary for object
  detection.
\newblock In {\em Proc. CVPR}, 2017.

\bibitem{WangZYB15}
Xinggang Wang, Zhuotun Zhu, Cong Yao, and Xiang Bai.
\newblock Relaxed multiple-instance {SVM} with application to object discovery.
\newblock In {\em Proc. ICCV}, 2015.

\bibitem{WeiFLCZY17}
Yunchao Wei, Jiashi Feng, Xiaodan Liang, Ming{-}Ming Cheng, Yao Zhao, and
  Shuicheng Yan.
\newblock Object region mining with adversarial erasing: {A} simple
  classification to semantic segmentation approach.
\newblock In {\em Proc. CVPR}, 2017.

\bibitem{ts2c}
Yunchao Wei, Zhiqiang Shen, Bowen Cheng, Honghui Shi, Jinjun Xiong, Jiashi
  Feng, and Thomas~S. Huang.
\newblock {TS2C:} tight box mining with surrounding segmentation context for
  weakly supervised object detection.
\newblock In {\em Proc. ECCV}, 2018.

\bibitem{reinforce}
Ronald~J. Williams.
\newblock Simple statistical gradient-following algorithms for connectionist
  reinforcement learning.
\newblock {\em Machine Learning}, 1992.

\bibitem{xiao-eccv2018}
Fanyi Xiao and Yong~Jae Lee.
\newblock Video object detection with an aligned spatial-temporal memory.
\newblock In {\em Proc. ECCV}, 2018.

\bibitem{Yang_2019_ICCV}
Ke Yang, Dongsheng Li, and Yong Dou.
\newblock Towards precise end-to-end weakly supervised object detection
  network.
\newblock In {\em Proc. ICCV}, 2019.

\bibitem{Zeng_2019_ICCV}
Zhaoyang Zeng, Bei Liu, Jianlong Fu, Hongyang Chao, and Lei Zhang.
\newblock {WSOD2}: Learning bottom-up and top-down objectness distillation for
  weakly-supervised object detection.
\newblock In {\em Proc. ICCV}, 2019.

\bibitem{zigzag}
Xiaopeng Zhang, Jiashi Feng, Hongkai Xiong, and Qi Tian.
\newblock Zigzag learning for weakly supervised object detection.
\newblock In {\em Proc. CVPR}, 2018.

\bibitem{Zhang_2018_CVPR}
Yongqiang Zhang, Yancheng Bai, Mingli Ding, Yongqiang Li, and Bernard Ghanem.
\newblock W2f: A weakly-supervised to fully-supervised framework for object
  detection.
\newblock In {\em Proc. CVPR}, 2018.

\bibitem{zhu17fgfa}
Xizhou Zhu, Yujie Wang, Jifeng Dai, Lu Yuan, and Yichen Wei.
\newblock Flow-guided feature aggregation for video object detection.
\newblock In {\em Proc. ICCV}, 2017.

\bibitem{eb}
C.~Lawrence Zitnick and Piotr Doll\'ar.
\newblock Edge boxes: Locating object proposals from edges.
\newblock In {\em Proc. ECCV}, 2014.

\bibitem{Zou_2019_ICCV}
Yang Zou, Zhiding Yu, Xiaofeng Liu, B.V.K.~Vijaya Kumar, and Jinsong Wang.
\newblock Confidence regularized self-training.
\newblock In {\em Proc. ICCV}, 2019.

\bibitem{zou2018unsupervised}
Yang Zou, Zhiding Yu, BVK Vijaya~Kumar, and Jinsong Wang.
\newblock Unsupervised domain adaptation for semantic segmentation via
  class-balanced self-training.
\newblock In {\em Proc. ECCV}, 2018.

\end{thebibliography}
}

\clearpage
% !TEX root = ./main.tex
{\centering \Large \textbf{Change Log}}
\begin{itemize}
    \item \textbf{v1}: ArXiv preprint. 
    \item \textbf{v2}: Additional implementation details (Appendix~\ref{app:implement}); Fixed a minor mistake in Fig.~\ref{fig:quali1}; Re-organized Appendix for better readability.
    \item \textbf{v3}: Fixed a typo in Alg.~\ref{alg:inst-or}.
\end{itemize}

\appendix
{\centering \Large \textbf{Appendix}}
\vspace{1em}

In this section, we provide:  (1) additional quantitative results on COCO; (2) per-class detection (AP) and correct localization (CorLoc) results on VOC; (3) additional qualitative results; (4) proposal statistics; (5) ablation study on the amount of proposals;  (6) implementation details and video demo of weakly supervised video object detection. Specifically, we show that our approach produces state-of-the-art results on COCO (see Tab.~\ref{table:coco-full}), outperforms all competing models on VOC 2007 and 2012 (see Tab.~\ref{table:per-cls-voc07} and Tab.~\ref{table:per-cls-voc12}). We also provide correct localization results in Tab.~\ref{table:per-cls-voc07-corloc} and Tab.~\ref{table:per-cls-voc12-corloc} for completeness and illustrate the necessity of the sequential batch back-propagation (introduced in  Sec.~\ref{sec:step-bp} of the main paper) in Tab.~\ref{table:prop} and Tab.~\ref{table:prop-exp}. Comprehensive visualizations are also  provided (Fig.~\ref{fig:quali_more_inst} to Fig.~\ref{fig:quali_more_vid}).

\section{Implementation Details}
\label{app:implement}
In this section, we provide additional implementation details for completeness. 
\subsection{Backbones}
\paragraph{VGG-16} We use the standard VGG-16 (without batch normalization) as backbone.
As shown in Fig.~\ref{fig:head}, the `Base' network contains all the convolutional layers before the fully-connected layers. Following~\cite{tang2017multiple}, we remove the last max-pooling layer, and replace the penultimate max-pooling layer and the subsequent convolutional layers with dilated convolutional (dilation=2) layers to increase the feature map  resolution. Standard RoI-pooling is used for computing region-level features. We use the fully-connected layers of VGG-16 except the last classifier layer as the `Neck'. After `Neck', the RoI features are projected to $f_w, g_w, \hat{s}_w, \hat{\mu}_w$ using 4 single fully-connected layers. 

\paragraph{ResNets} We use the \texttt{ResNet-50/101-C4} variant from Detectron code repository~\cite{Detectron2018}. Convolutional layers of the first 4 ResNet stages (C1-C4) are used as `Base' and the last stage (C5) is used as `Neck'. Standard RoI-pooling is used, and RoI features are projected using linear layers.

\subsection{Concrete DropBlock}
Concrete DropBlock is implemented as a standard residual block as in ResNets. It takes as input the RoI features and output a 1 channel heatmap $p_\theta(r)$. On the skip connection we use $1\times 1$ convolution to reduce feature channels. We then generate the hard mask $M_\theta(r)$ using Gumbel-softmax, and the structured dropout region as in DropBlock~\cite{dropblock}.

\subsection{Student Blocks}
Following~\cite{tang2017multiple}, we stack 3 student blocks. During training, student block $N$ generates pseudo labels for the next student block $N+1$. During testing, we average the predictions of all student blocks as final results.

\subsection{Training}
Our code is implemented in PyTorch and all the experiments are conducted on single 8-GPU (NVIDIA V100) machine. SGD is used for optimization with weight decay 0.0001 and momentum 0.9. The batch size and initial learning rate is set to 8 and 0.01 on VOC 2007; 16 and 0.02 on VOC 2012. On both datasets we train the model for 30k iterations and decay the learning rate by 0.1 at 20k and 26k steps. On COCO, we train the model for total 130k iterations and decay the learning rate at 90k and 120k steps with batch size 8 and initial learning rate 0.01. We use Selective-Search (SS)~\cite{ss} for VOC datasets and MCG~\cite{mcg} for COCO.

\subsection{Data Augmentation \& Inference}
Multi-scale inputs (480, 576, 688, 864, 1000, 1200) are used during both  training and testing following ~\cite{tang2017multiple,KantorovOCL16} and the longest image side to set to less than 2000.
At test time, the scores are averaged over all scales and their horizontal flips.

\section{Additional quantitative results on COCO}
\label{app:coco}
In Tab.~\ref{table:coco-full}, we report  quantitative results at different thresholds and scales on COCO for different models. The reported metrics include: Average Prevision ($AP$)  over multiple IoU thresholds (.50 : .05 : .95), at IoU threshold 50\% and 75\% ($AP^{50}$, $AP^{75}$), and for small, medium and large objects ($AP^{\text{s}}$, $AP^{\text{m}}$, $AP^{\text{l}}$); and Average Recall ($AR$) over multiple IoU values (.50 : .05 : .95), given 1, 10 and 100 detections per image ($AR^{1}, AR^{10}, AR^{100}$); and for small, medium and large objects ($AR^{\text{s}}$, $AR^{\text{m}}$, $AR^{\text{l}}$). The results in Tab.~\ref{table:coco-full} show that object size is a significant factor that influences the detection accuracy. The detector tends to perform better on large objects rather than smaller ones. 

\begin{table*}[h]
\centering\footnotesize{
\begin{tabular}{c c c| c c c c c c |c  c c c c c }
\specialrule{.15em}{.05em}{.05em}
Train & Test & Model & $AP$ & $AP^{50}$ & $AP^{75}$ & $AP^{\text{s}}$  &  $AP^{\text{m}}$  &  $AP^{\text{l}}$  & $AR^{1}$ &   $AR^{10}$ &  $AR^{100}$ & $AR^{\text{s}}$  &  $AR^{\text{m}}$  &  $AR^{\text{l}}$ \\ \hline
2014 Train & 2014 Val & VGG16 & 11.4 & 24.3 & 9.4 & 3.6 & 12.2 & 17.6 & 13.5 & 22.6 & 23.9 & 8.5 & 25.4 & 38.3 \\
2014 Train & 2014 Val & R50-C4  & 12.6 & 26.1 & 10.8 & 3.7 & 13.3 & 19.9 & 14.8 & 23.7 & 24.7 & 8.4 & 25.1 & 41.8\\
2014 Train & 2014 Val & R101-C4 & 13.0 & 26.3 & 11.4 & 3.5 & 13.7 & 20.4 & 15.4 & 23.4 & 24.6 & 8.5 & 24.6 & 40.9\\
\hline
2017 Train & minival & VGG16 & 12.4 & 25.8 & 10.5 & 3.9 & 13.8 & 19.9 & 14.3 & 23.3 & 24.6 & 9.7 & 26.6 & 39.6\\
\hline
2014 Train & Test-Dev & VGG16 & 12.1 & 24.8 & 10.2 & 4.1 & 13.0 & 18.3 & 13.5 & 25.5 & 29.0 & 9.6 & 30.0 & 46.7 \\ 
\specialrule{.15em}{.05em}{.05em}
\end{tabular}}
\caption{Single model detection results on COCO.}
\label{table:coco-full}
\end{table*}

\section{Additional results on VOC}
\label{app:voc}
\subsection{Per-class detection results}

\begin{table*}[t]
\centering
\resizebox{\textwidth}{!}{
{
\begin{tabular}{c | c | c c c c c c c c c c c c c c c c c c c c | c}
\specialrule{.15em}{.05em}{.05em}
Methods & Proposal & Aero & Bike & Bird & Boat & Bottle & Bus & Car & Cat & Chair & Cow & Table & Dog & Horse & Motor & Person & Plant & Sheep & Sofa & Train & ~~TV~~ & ~~AP~~ \\
\hline
Fast R-CNN & SS  & 73.4 &  77.0 &  63.4 &  45.4 &  44.6 &  75.1 &  78.1  & 79.8 &  40.5 &  73.7 &  62.2 &  79.4  & 78.1 &  73.1 &  64.2  & 35.6  & 66.8 &  67.2 &  70.4  & 71.1 & 66.0  \\
Faster R-CNN  & RPN  & 70.0 &  80.6  & 70.1 &  57.3  & 49.9  & 78.2  & 80.4  & 82.0  &  52.2 &  75.3  & 67.2  & 80.3  & 79.8  & 75.0  & 76.3  & 39.1  & 68.3 &  67.3  & 81.1 &  67.6  & \textbf{69.9}  \\
\hline
Cinbis~\cite{CinbisVS14}  & SS & 35.8 &  40.6 &  8.1 &  7.6 &  3.1 &  35.9 &  41.8 &  16.8 &  1.4 &  23.0 &  4.9 &  14.1 &  31.9 &  41.9 &  19.3 &  11.1 &  27.6 &  12.1 &  31.0 &  40.6 & 22.4 \\
Bilen~\cite{BilenPT15} & SS & 46.2 &  46.9 &  24.1 &  16.4 &  12.2 &  42.2 &  47.1 &  35.2 &  7.8 &  28.3 &  12.7 &  21.5 &  30.1 &  42.4 &  7.8 &  20.0 &  26.8 &  20.8 &  35.8 &  29.6 & 27.7 \\
Wang~\cite{WangRHT14} & SS  & 48.8 &  41.0 &  23.6 &  12.1 &  11.1 &  42.7 &  40.9 &  35.5 &  11.1 &  36.6 &  18.4 &  35.3 &  34.8 &  51.3 &  17.2 &  17.4 &  26.8 &  32.8 &  35.1 &  45.6 & 30.9 \\
Li~\cite{LiHLW016} & EB  & 54.5 & 47.4 & 41.3 & 20.8 & 17.7 & 51.9 & 63.5 & 46.1 & 21.8 & 57.1 & 22.1 & 34.4 & 50.5 & 61.8 & 16.2 & \textbf{29.9} & 40.7 & 15.9 & 55.3 & 40.2 & 39.5 \\
WSDDN~\cite{Bilen16} & EB  &39.4 & 50.1 & 31.5 & 16.3 & 12.6 & 64.5 & 42.8 & 42.6 & 10.1 & 35.7 & 24.9 & 38.2 & 34.4 & 55.6 & 9.4 & 14.7 & 30.2 & 40.7 & 54.7 & 46.9 & 34.8 \\
Teh~\cite{TehRW16} & EB  & 48.8 & 45.9 & 37.4 & 26.9 & 9.2 & 50.7 & 43.4 & 43.6 & 10.6 & 35.9 & 27.0 & 38.6 & 48.5 & 43.8 & 24.7 & 12.1 & 29.0 & 23.2 & 48.8 & 41.9 & 34.5 \\
ContextLocNet~\cite{KantorovOCL16}  & SS  & 57.1 & 52.0 & 31.5 & 7.6 & 11.5 & 55.0 & 53.1 & 34.1 & 1.7 & 33.1 & 49.2 & 42.0 & 47.3 & 56.6 & 15.3 & 12.8 & 24.8 & 48.9 & 44.4 & 47.8 & 36.3 \\
OICR~\cite{tang2017multiple}   & SS & 58.0 & 62.4 & 31.1 & 19.4 & 13.0 & 65.1 & 62.2 & 28.4 & 24.8 & 44.7 & 30.6 & 25.3 & 37.8 & 65.5 & 15.7 & 24.1 & 41.7 & 46.9 & 64.3 & 62.6 & 41.2 \\
Jie~\cite{JieWJFL17} & ?  & 52.2 & 47.1 & 35.0 & 26.7 & 15.4 & 61.3 & 66.0 & 54.3 & 3.0 & 53.6 & 24.7 & 43.6 & 48.4 & 65.8 & 6.6 & 18.8 & 51.9 & 43.6 & 53.6 & 62.4 & 41.7 \\
Diba~\cite{DibaSPPG17} & EB &  49.5 & 60.6 & 38.6 & 29.2 & 16.2 & 70.8 & 56.9 & 42.5 & 10.9 & 44.1 & 29.9 & 42.2 & 47.9 & 64.1 & 13.8 & 23.5 & 45.9 & 54.1 & 60.8 & 54.5 & 42.8 \\
PCL~\cite{tang2018pcl}  & SS & 54.4 & 69.0 & 39.3 & 19.2 & 15.7 & 62.9 & 64.4 & 30.0 & 25.1 & 52.5 & 44.4 & 19.6 & 39.3 & 67.7 & 17.8 & 22.9 & 46.6 & 57.5 & 58.6 & 63.0 & 43.5 \\
Wei~\cite{ts2c} & SS & 59.3 & 57.5 & 43.7 & 27.3 & 13.5 & 63.9 & 61.7 & 59.9 & 24.1 & 46.9 & 36.7 & 45.6 & 39.9 & 62.6 & 10.3 & 23.6 & 41.7 & 52.4 & 58.7 & 56.6 & 44.3 \\
Tang~\cite{TangWWYLHY18}  & SS & 57.9 & 70.5 & 37.8 & 5.7 & 21.0 & 66.1 & 69.2 & 59.4 & 3.4 & 57.1 & \textbf{57.3} & 35.2 & 64.2 & 68.6 & 32.8 & 28.6 & 50.8 & 49.5 & 41.1 & 30.0 & 45.3 \\
Shen~\cite{Shen_2019_CVPR} & SS & 52.0 & 64.5 & 45.5 & 26.7 & 27.9 & 60.5 & 47.8 & 59.7 & 13.0 & 50.4 & 46.4 & 56.3 & 49.6 & 60.7 & 25.4 & 28.2 & 50.0 & 51.4 & 66.5 & 29.7 & 45.6 \\
Wan~\cite{Wan_2018_CVPR}  & SS & 55.6 & 66.9 & 34.2 & 29.1 & 16.4 & 68.8 & 68.1 & 43.0 & 25.0 & 65.6 & 45.3 & 53.2 & 49.6 & 68.6 & 2.0 & 25.4 & 52.5 & 56.8 & 62.1 & 57.1 & 47.3 \\
SDCN~\cite{Li_2019_ICCV} & SS & 59.4 & 71.5 & 38.9 & 32.2 & 21.5 & 67.7 & 64.5 & \textbf{68.9} & 20.4 & 49.2 & 47.6 & 60.9 & 55.9 & 67.4 & 31.2 & 22.9 & 45.0 & 53.2 & 60.9 & 64.4 & 50.2 \\
C-MIL~\cite{c-mil} & SS & 62.5 & 58.4 & 49.5 & 32.1 & 19.8 & 70.5 & 66.1 & 63.4 & 20.0 & 60.5 & 52.9 & 53.5 & 57.4 & 68.9 & 8.4 & 24.6 & 51.8 & 58.7 & 66.7 & 63.6 & 50.5 \\
Yang~\cite{Yang_2019_ICCV} & SS & 57.6 & 70.8 & 50.7 & 28.3 & 27.2 & 72.5 & 69.1 & 65.0 & 26.9 & 64.5 & 47.4 & 47.7  & 53.5 & 66.9 & 13.7 & 29.3 & 56.0 & 54.9 & 63.4 & 65.2 & 51.5 \\
C-MIDN~\cite{Gao_2019_ICCV} & SS & 53.3 & 71.5 & 49.8 & 26.1 & 20.3 & 70.3 & 69.9 & 68.3 & 28.7 & 65.3 & 45.1 & \textbf{64.6} & 58.0 & 71.2 & 20.0 & 27.5 & 54.9 & 54.9 & \textbf{69.4} & 63.5 & 52.6 \\
Arun~\cite{Arun_2019} & SS & 66.7 & 69.5 & 52.8 & 31.4 & 24.7 & \textbf{74.5} & 74.1 & 67.3 & 14.6 & 53.0 & 46.1 & 52.9 & 69.9 & 70.8 & 18.5 & 28.4 & 54.6 & \textbf{60.7} & 67.1 & 60.4 & 52.9 \\
WSOD2~\cite{Zeng_2019_ICCV}  & SS & 65.1 & 64.8 & \textbf{57.2} & \textbf{39.2} & 24.3 & 69.8 & 66.2 & 61.0 & 29.8 & 64.6 & 42.5 & 60.1 & \textbf{71.2} & 70.7 & 21.9 & 28.1 & 58.6 & 59.7 & 52.2 & 64.8 & 53.6 \\
\hline
Ours  & SS & \textbf{68.8} & \textbf{77.7} & 57.0 & 27.7 & \textbf{28.9} & 69.1 & \textbf{74.5} & 67.0 & \textbf{32.1} & \textbf{73.2} & 48.1 & 45.2 & 54.4 & \textbf{73.7} & \textbf{35.0} & 29.3 & \textbf{64.1} & 53.8 & 65.3 & \textbf{65.2} & \textbf{54.9}  \\
\specialrule{.15em}{.05em}{.05em}
\end{tabular}
}}
\caption{Single model per-class detection results using VGG16 on PASCAL VOC 2007.}
\label{table:per-cls-voc07}
\vspace{1em}
\centering
\resizebox{\textwidth}{!}{
{
\begin{tabular}{c | c | c c c c c c c c c c c c c c c c c c c c | c}
\specialrule{.15em}{.05em}{.05em}
Methods & Proposal & Aero & Bike & Bird & Boat & Bottle & Bus & Car & Cat & Chair & Cow & Table & Dog & Horse & Motor & Person & Plant & Sheep & Sofa & Train & ~~TV~~ & ~~AP~~ \\
\hline
Fast R-CNN    & SS  & 80.3 & 74.7 & 66.9 & 46.9 & 37.7 & 73.9 & 68.6 & 87.7 & 41.7 & 71.1 & 51.1 & 86.0 & 77.8 & 79.8 & 69.8 & 32.1 & 65.5 & 63.8 & 76.4 & 61.7 & 65.7\\
Faster R-CNN  & RPN & 82.3 & 76.4 & 71.0 & 48.4 & 45.2 & 72.1 & 72.3 & 87.3 & 42.2 & 73.7 & 50.0 & 86.8 & 78.7 & 78.4 & 77.4 & 34.5 & 70.1 & 57.1 & 77.1 & 58.9 & 67.0\\
\hline
Li~\cite{LiHLW016} & EB  & 62.9 & 55.5 & 43.7 & 14.9 & 13.6 & 57.7 & 52.4 & 50.9 & 13.3 & 45.4 & 4.0 & 30.2 & 55.6 & 67.0 & 3.8 & 23.1 & 39.4 & 5.5 & 50.7 & 29.3 & 35.9 \\
ContextLocNet~\cite{KantorovOCL16} & SS & 64.0 & 54.9 & 36.4 & 8.1 & 12.6 & 53.1 & 40.5 & 28.4 & 6.6 & 35.3 & \textbf{34.4} & 49.1 & 42.6 & 62.4 & \textbf{19.8} & 15.2 & 27.0 & 33.1 & 33.0 & 50.0 & 35.3\\
OICR~\cite{tang2017multiple}   & SS & 67.7 & 61.2 & 41.5 & 25.6 & 22.2 & 54.6 & 49.7 & 25.4 & 19.9 & 47.0 & 18.1 & 26.0 & 38.9 & 67.7 & 2.0 & 22.6 & 41.1 & 34.3 & 37.9 & 55.3 & 37.9 \\
Jie~\cite{JieWJFL17}      & ?  & 60.8 & 54.2 & 34.1 & 14.9 & 13.1 & 54.3 & 53.4 & 58.6 & 3.7 & 53.1 & 8.3 & 43.4 & 49.8 & 69.2 & 4.1 & 17.5 & 43.8 & 25.6 & 55.0 & 50.1 & 38.3 \\
Diba~\cite{DibaSPPG17}    & EB &-&-&-&-&-&-&-&-&-&-&-&-&-&-&-&-&-&-&-&-& 37.9 \\
Shen~\cite{Shen_2019_CVPR}  & SS &-&-&-&-&-&-&-&-&-&-&-&-&-&-&-&-&-&-&-&-& 39.1\\
PCL~\cite{tang2018pcl}    & SS & 58.2 & 66.0 & 41.8 & 24.8 & 27.2 & 55.7 & 55.2 & 28.5 & 16.6 & 51.0 & 17.5 & 28.6 & 49.7 & 70.5 & 7.1 & 25.7 & 47.5 & 36.6 & 44.1 & 59.2 & 40.6\\
Wei~\cite{ts2c} & SS & 67.4 & 57.0 & 37.7 & 23.7 & 15.2 & 56.9 & 49.1 & 64.8 & 15.1 & 39.4 & 19.3 & 48.4 & 44.5 & 67.2 & 2.1 & 23.3 & 35.1 & 40.2 & 46.6 & 45.8 & 40.0\\
Tang~\cite{TangWWYLHY18}  & SS &-&-&-&-&-&-&-&-&-&-&-&-&-&-&-&-&-&-&-&-& 40.8\\
Wan~\cite{Wan_2018_CVPR}  & SS &-&-&-&-&-&-&-&-&-&-&-&-&-&-&-&-&-&-&-&-& 42.4\\
SDCN~\cite{Li_2019_ICCV}  & SS &-&-&-&-&-&-&-&-&-&-&-&-&-&-&-&-&-&-&-&-& 43.5 \\
Yang~\cite{Yang_2019_ICCV} & SS & 64.7 & 66.3 & 46.8 & 28.5 & 28.4 & 59.8 & 58.6 & 70.9 & 13.8 & 55.0 & 15.7 & 60.5 &  \textbf{63.9} & 69.2 & 8.7 & 23.8 & 44.7 &  \textbf{52.7} & 41.5 & 62.6 & 46.8 \\
C-MIL~\cite{c-mil} & SS &-&-&-&-&-&-&-&-&-&-&-&-&-&-&-&-&-&-&-&-& 46.7 \\
WSOD2~\cite{Zeng_2019_ICCV}  & SS &-&-&-&-&-&-&-&-&-&-&-&-&-&-&-&-&-&-&-&-& 47.2\\
Arun~\cite{Arun_2019} & SS &-&-&-&-&-&-&-&-&-&-&-&-&-&-&-&-&-&-&-&-& 48.4 \\
C-MIDN~\cite{Gao_2019_ICCV} & SS & 72.9 & 68.9 & 53.9 & 25.3 & 29.7 & 60.9 & 56.0 &  \textbf{78.3} & 23.0 & 57.8 & 25.7 &  \textbf{73.0} & 63.5 & 73.7 & 13.1 & 28.7 & 51.5 & 35.0 & 56.1 & 57.5 & 50.2 \\
\hline
Ours\footnotemark  & SS & \textbf{78.3} & \textbf{73.9} & \textbf{56.5} & \textbf{30.4} & \textbf{37.4} & \textbf{64.2} & \textbf{59.3} & 60.3 & \textbf{26.6} & \textbf{66.8} & 25.0 & 55.0 & 61.8 & \textbf{79.3} & 14.5 & \textbf{30.3} & \textbf{61.5} & 40.7 & \textbf{56.4} & \textbf{63.5} & \textbf{52.1}   \\
\specialrule{.15em}{.05em}{.05em}
\end{tabular}
}}
\caption{Single model per-class detection results using VGG16 on PASCAL VOC 2012.}
\label{table:per-cls-voc12}
\end{table*}

In Tab.~\ref{table:per-cls-voc07} and Tab.~\ref{table:per-cls-voc12}, we report the per-class detection APs on the test sets of both VOC 2007 and 2012. Compared to other WSOD methods we observe: (1) Our method outperforms all others on most categories (10 classes on VOC 2007, 14 classes on VOC 2012). (2) The classes that are hard for our approach (\eg, boat, plant, and chair) are also challenging for other methods. This suggests that these categories are essentially hard examples for WSOD methods, for which a certain amount of strong supervision might still be needed.

Compared to supervised models (Fast R-CNN, Faster R-CNN) we note: (1) Our weakly supervised model performs competitively for classes such as: airplane, bicycle, bus, car, cow, motorbike, sheep, tv-monitor, where the performance gap is usually less than 10\% AP. Our model sometimes even outperforms supervised models on categories that are considered relatively easy with small intra-class difference (bicycle and motorbike in VOC 2007, motorbike and tv-monitor in VOC 2012). (2) For classes like boat, chair, dinning table, person, all WSOD methods are significantly worse than supervised methods. This is likely due to a large intra-class variation. WSOD methods fail to capture the consistent patterns of these classes.

\subsection{Per-class correct localization results}
\begin{table*}[t]
\centering
\resizebox{\textwidth}{!}{
{
\begin{tabular}{c | c | c c c c c c c c c c c c c c c c c c c c | c}
\specialrule{.15em}{.05em}{.05em}
Methods & Proposal & Aero & Bike & Bird & Boat & Bottle & Bus & Car & Cat & Chair & Cow & Table & Dog & Horse & Motor & Person & Plant & Sheep & Sofa & Train & ~~TV~~ & CorLoc \\
\hline
Cinbis~\cite{CinbisVS14}& SS  & 56.6 & 58.3 & 28.4 & 20.7 & 6.8 & 54.9 & 69.1 & 20.8 & 9.2 & 50.5 & 10.2 & 29.0 & 58.0 & 64.9 & 36.7 & 18.7 & 56.5 & 13.2 & 54.9 & 59.4 & 38.8 \\
Bilen~\cite{BilenPT15} & SS  & 66.4 & 59.3 & 42.7 & 20.4 & 21.3 & 63.4 & 74.3 & 59.6 & 21.1 & 58.2 & 14.0 & 38.5 & 49.5 & 60.0 & 19.8 & 39.2 & 41.7 & 30.1 & 50.2 & 44.1 & 43.7 \\
Wang~\cite{WangRHT14} & SS   & 80.1 & 63.9 & 51.5 & 14.9 & 21.0 & 55.7 & 74.2 & 43.5 & 26.2 & 53.4 & 16.3 & 56.7 & 58.3 & 69.5 & 14.1 & 38.3 & 58.8 & 47.2 & 49.1 & 60.9 & 48.5  \\
Li~\cite{LiHLW016} & EB &  78.2 & 67.1 & 61.8 & 38.1 & 36.1 & 61.8 & 78.8 & 55.2 & 28.5 & 68.8 & 18.5 & 49.2 & 64.1 & 73.5 & 21.4 & 47.4 & 64.6 & 22.3 & 60.9 & 52.3 & 52.4 \\
WSDDN~\cite{Bilen16} & EB  & 65.1 & 58.8 & 58.5 & 33.1 & 39.8 & 68.3 & 60.2 & 59.6 & 34.8 & 64.5 & 30.5 & 43.0 & 56.8 & 82.4 & 25.5 & 41.6 & 61.5 & 55.9 & 65.9 & 63.7 & 53.5 \\
Teh~\cite{TehRW16} & EB  & 84.0 & 64.6 & 70.0 & \textbf{62.4} & 25.8 & 80.6 & 73.9 & 71.5 & 35.7 & 81.6 & 46.5 & 71.3 & 79.1 & 78.8 & 56.7 & 34.3 & 69.8 & 56.7 &77.0 & 72.7 & 64.6 \\
ContextLocNet~\cite{KantorovOCL16}  & SS & 83.3 & 68.6 & 54.7 & 23.4 & 18.3 & 73.6 & 74.1 & 54.1 & 8.6 & 65.1 & 47.1 & 59.5 & 67.0 & 83.5 & 35.3 & 39.9 & 67.0 & 49.7 & 63.5 & 65.2 & 55.1\\
OICR~\cite{tang2017multiple}   & SS & 81.7 & 80.4 & 48.7 & 49.5 & 32.8 & 81.7 & 85.4 & 40.1 & 40.6 & 79.5 & 35.7 & 33.7 & 60.5 & 88.8 & 21.8 & 57.9 & 76.3 & 59.9 & 75.3 & \textbf{81.4} & 60.6 \\
Jie~\cite{JieWJFL17}      & ?  & 72.7 & 55.3 & 53.0 & 27.8 & 35.2 & 68.6 & 81.9 & 60.7 & 11.6 & 71.6 & 29.7 & 54.3 & 64.3 & 88.2 & 22.2 & 53.7 & 72.2 & 52.6 & 68.9 & 75.5 & 56.1\\
Diba~\cite{DibaSPPG17}    & EB  & 83.9 & 72.8 & 64.5 & 44.1 & 40.1 & 65.7 & 82.5 & 58.9 & 33.7 & 72.5 & 25.6 & 53.7 & 67.4 & 77.4 & 26.8 & 49.1 & 68.1 & 27.9 & 64.5 & 55.7 & 56.7 \\
Wei~\cite{ts2c}           & SS  & 84.2 & 74.1 & 61.3 & 52.1 & 32.1 & 76.7 & 82.9 & 66.6 & 42.3 & 70.6 & 39.5 & 57.0 & 61.2 & 88.4 & 9.3 & 54.6 & 72.2 & 60.0 & 65.0 & 70.3 & 61.0 \\
Wan~\cite{Wan_2018_CVPR}  & SS  &-&-&-&-&-&-&-&-&-&-&-&-&-&-&-&-&-&-&-&- & 61.4\\
PCL~\cite{tang2018pcl}    & SS  & 79.6 & 85.5 & 62.2 & 47.9 & 37.0 & 83.8 & 83.4 & 43.0 & 38.3 & 80.1 & 50.6 & 30.9 & 57.8 & 90.8 & 27.0 & 58.2 & 75.3 & \textbf{68.5} & 75.7 & 78.9 & 62.7 \\
Tang~\cite{TangWWYLHY18}  & SS  & 77.5 & 81.2 & 55.3 & 19.7 & 44.3 & 80.2 & 86.6 & 69.5 & 10.1 & 87.7 & \textbf{68.4} & 52.1 & 84.4 & 91.6 & \textbf{57.4} & \textbf{63.4} & 77.3 & 58.1 & 57.0 & 53.8 & 63.8 \\
Li~\cite{Li_2019_ICCV} & SS & 85.0 & 83.9 & 58.9 & 59.6 & 43.1 & 79.7 & 85.2 & 77.9 & 31.3 & 78.1 & 50.6 & 75.6 & 76.2 & 88.4 & 49.7 & 56.4 & 73.2 & 62.6 & 77.2 & 79.9 & 68.6 \\
Shen~\cite{Shen_2019_CVPR} & SS & 82.9 & 74.0 & 73.4 & 47.1 & \textbf{60.9} & 80.4 & 77.5 & \textbf{78.8} & 18.6 & 70.0 & 56.7 & 67.0 & 64.5 & 84.0 & 47.0 & 50.1 & 71.9 & 57.6 & \textbf{83.3} & 43.5 & 64.5 \\
C-MIL~\cite{c-mil} & SS&-&-&-&-&-&-&-&-&-&-&-&-&-&-&-&-&-&-&-&-& 65.0 \\
Yang~\cite{Yang_2019_ICCV} & SS & 80.0 & 83.9 & 74.2 & 53.2 & 48.5 & 82.7 & 86.2 & 69.5 & 39.3 & 82.9 & 53.6 & 61.4& 72.4 & 91.2 & 22.4 & 57.5 & \textbf{83.5} & 64.8 & 75.7 & 77.1 & 68.0 \\
WSOD2~\cite{Zeng_2019_ICCV} & SS & 87.1 & 80.0 & 74.8 & 60.1 & 36.6 & 79.2 & 83.8 & 70.6 & 43.5 & \textbf{88.4} & 46.0 & \textbf{74.7} & 87.4 & 90.8 & 44.2 & 52.4 & 81.4 & 61.8 & 67.7 & 79.9 & 69.5\\
Arun~\cite{Arun_2019} & SS & \textbf{88.6} & \textbf{86.3} & 71.8 & 53.4 & 51.2 & \textbf{87.6} & \textbf{89.0} & 65.3 & 33.2 & 86.6 & 58.8 & 65.9 & \textbf{87.7} & \textbf{93.3} & 30.9 & 58.9 & 83.4 & 67.8 & 78.7 & 80.2 & \textbf{70.9} \\
\hline
Ours & SS & 87.5 & 82.4 & \textbf{76.0} & 58.0 & 44.7 & 82.2 & 87.5 & 71.2 & \textbf{49.1} & 81.5 & 51.7 & 53.3 & 71.4 & 92.8 & 38.2 & 52.8 &79.4 & 61.0 & 78.3 & 76.0  & 68.8 \\
\specialrule{.15em}{.05em}{.05em}
\end{tabular}
}}
\caption{Single model per-class correct localization (CorLoc) results using VGG16 on PASCAL VOC 2007.}
\label{table:per-cls-voc07-corloc}
\end{table*}

\begin{table*}[tbh]
\centering
\resizebox{\textwidth}{!}{
{
\begin{tabular}{c | c | c c c c c c c c c c c c c c c c c c c c | c}
\specialrule{.15em}{.05em}{.05em}
Methods & Proposal & Aero & Bike & Bird & Boat & Bottle & Bus & Car & Cat & Chair & Cow & Table & Dog & Horse & Motor & Person & Plant & Sheep & Sofa & Train & ~~TV~~ & CorLoc \\
\hline
Li~\cite{LiHLW016} & EB  &-&-&-&-&-&-&-&-&-&-&-&-&-&-&-&-&-&-&-&- & 29.1 \\
ContextLocNet~\cite{KantorovOCL16} & SS &78.3 & 70.8 & 52.5 & 34.7 & 36.6 & 80.0 & 58.7 & 38.6 & 27.7 & 71.2 & 32.3 & 48.7 & 76.2 & 77.4 & 16.0 & 48.4 & 69.9 & 47.5 & 66.9 & 62.9 & 54.8 \\
OICR~\cite{tang2017multiple}   & SS  &-&-&-&-&-&-&-&-&-&-&-&-&-&-&-&-&-&-&-&-& 62.1 \\
Jie~\cite{JieWJFL17}      & ?  & 82.4 & 68.1 & 54.5 & 38.9 & 35.9 & 84.7 & 73.1 & 64.8 & 17.1 & 78.3 & 22.5 & 57.0 & 70.8 & 86.6 & 18.7 & 49.7 & 80.7 & 45.3 & 70.1 & 77.3 & 58.8 \\
PCL~\cite{tang2018pcl}    & SS & 77.2 & 83.0 & 62.1 & 55.0 & 49.3 & 83.0 & 75.8 & 37.7 & 43.2 & 81.6 & 46.8 & 42.9 & 73.3 & 90.3 & 21.4 & 56.7 & 84.4 & 55.0 & 62.9 & 82.5 & 63.2 \\
Wei~\cite{ts2c} & SS & 79.1 & 83.9 & 64.6 & 50.6 & 37.8 & 87.4 & 74.0 &  74.1 & 40.4 & 80.6 & 42.6 & 53.6 & 66.5 & 88.8 & 18.8 & 54.9 & 80.4 &  60.4 & 70.7 & 79.3 & 64.4 \\
Shen~\cite{Shen_2019_CVPR} & SS &-&-&-&-&-&-&-&-&-&-&-&-&-&-&-&-&-&-&-&-& 63.5 \\
Tang~\cite{TangWWYLHY18}  & SS & 85.5 & 60.8 & 62.5 & 36.6 & 53.8 & 82.1 &  \textbf{80.1} & 48.2 & 14.9 & 87.7 & \textbf{68.5} &  60.7 &  85.7 & 89.2 &  \textbf{62.9} &  \textbf{62.1} & 87.1 & 54.0 & 45.1 & 70.6 & 64.9 \\
Li~\cite{Li_2019_ICCV} & SS &-&-&-&-&-&-&-&-&-&-&-&-&-&-&-&-&-&-&-&-& 67.9 \\
C-MIL~\cite{c-mil} & SS &-&-&-&-&-&-&-&-&-&-&-&-&-&-&-&-&-&-&-&-& 67.4 \\
Yang~\cite{Yang_2019_ICCV}  & SS & 82.4 & 83.7 & \textbf{72.4} & \textbf{57.9} & 52.9 & 86.5 & 78.2 & \textbf{78.6} & 40.1 & 86.4 & 37.9 & \textbf{67.9} & \textbf{87.6} & 90.5 & 25.6 & 53.9 & 85.0 & \textbf{71.9} & 66.2& 84.7 & 69.5 \\
Arun~\cite{Arun_2019} & SS &-&-&-&-&-&-&-&-&-&-&-&-&-&-&-&-&-&-&-&-& 69.5 \\
WSOD2~\cite{Zeng_2019_ICCV} & SS &-&-&-&-&-&-&-&-&-&-&-&-&-&-&-&-&-&-&-&-& \textbf{71.9} \\
\hline
Ours  & SS & \textbf{91.7} & \textbf{85.6} & 71.7 & 56.6 & \textbf{55.6} & \textbf{88.6} & 77.3 & 63.4 & \textbf{53.6} & \textbf{90.0} & 51.6 & 62.6 & 79.3 & \textbf{94.2} & 32.7 & 58.8 & \textbf{90.5} & 57.7 & \textbf{70.9} & \textbf{85.7} & 70.9 \\
\specialrule{.15em}{.05em}{.05em}
\end{tabular}
}}
\caption{Single model per-class correct localization (CorLoc) results using VGG16 on PASCAL VOC 2012.}
\label{table:per-cls-voc12-corloc}
\end{table*}

In Tab.~\ref{table:per-cls-voc07-corloc} and Tab.~\ref{table:per-cls-voc12-corloc}, we report the per-class correct localization (CorLoc) results on the trainval sets of both VOC 2007 and VOC 2012. Consistent with prior work~\cite{Bilen16, tang2017multiple, c-mil, Zeng_2019_ICCV, Zhang_2018_CVPR, Arun_2019} this metric is computed on the training set. Thus it does not reflect the true performance of the detection models and has not been widely adopted by supervised methods~\cite{fastrcnn, ren16faster, he2017maskrcnn}. For WSOD approaches, it serves as an indicator of the `over-fitting' behavior. Compared with previous state-of-the-art, our method achieves the third best result on VOC 2007, winning on 2 categories. We also achieve the second best performance on VOC 2012 and win on 19 categories. We find that: (1) Our model performs well for  classes like: airplane, bicycle, bottle, bus, motorbike, sheep, tv-monitor. This observation aligns very well with the detection results. (2) The best performing methods differ across  classes, which suggest that  methods could potentially be ensembled for further improvements. 

\footnotetext{http://host.robots.ox.ac.uk:8080/anonymous/DCJ5GA.html}

\section{Additional qualitative results}
\label{app:fig}
\subsection{Results on static-image datasets}
\begin{figure*}[t]
\centering
\includegraphics[width=0.98\textwidth]{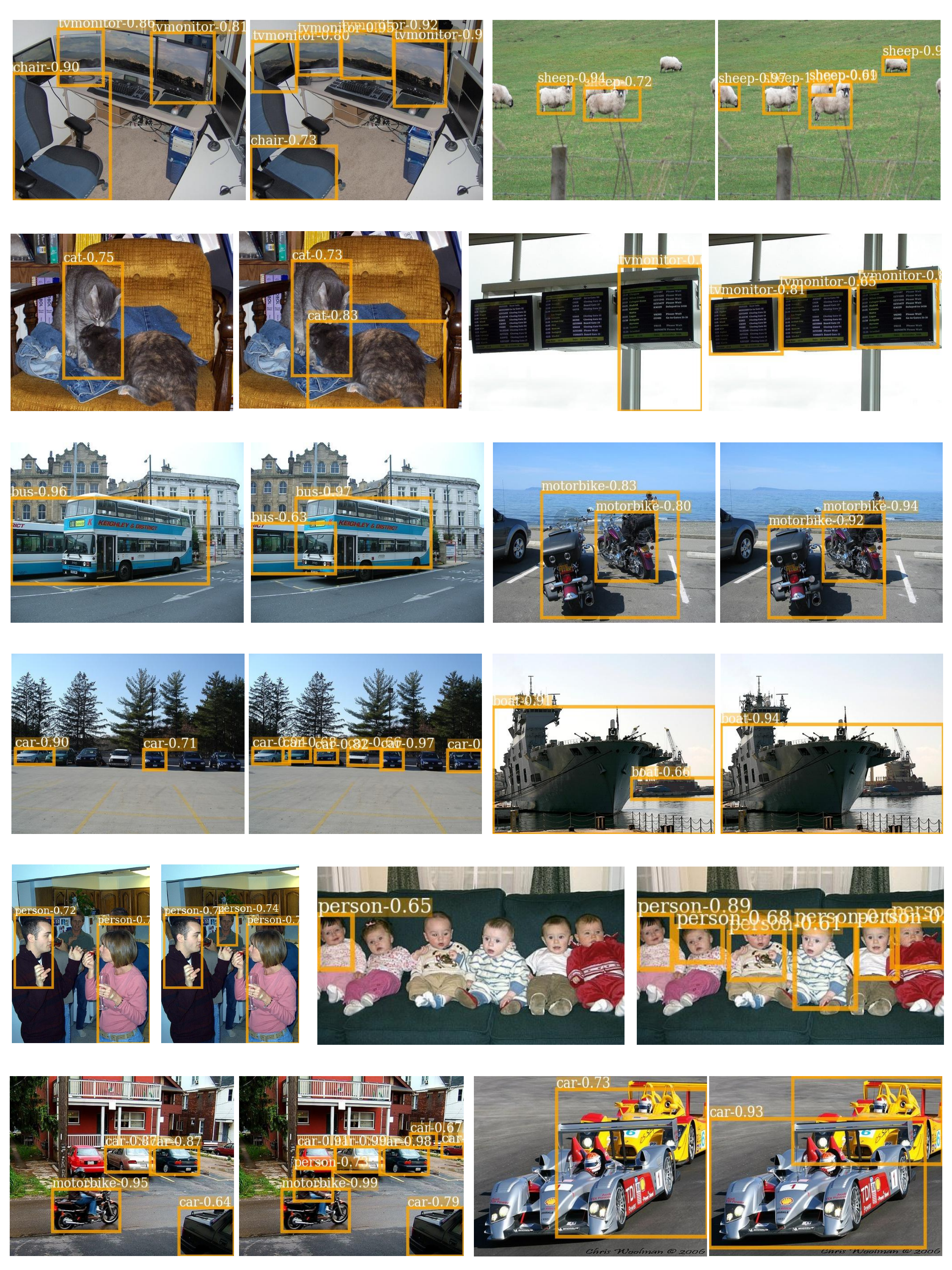}
\caption{Examples that highlight cases of `Instance Ambiguity'. For every pair: baseline (left) and our model (right).}
\label{fig:quali_more_inst}
\end{figure*}

\begin{figure*}[t]
\centering
\includegraphics[width=0.98\textwidth]{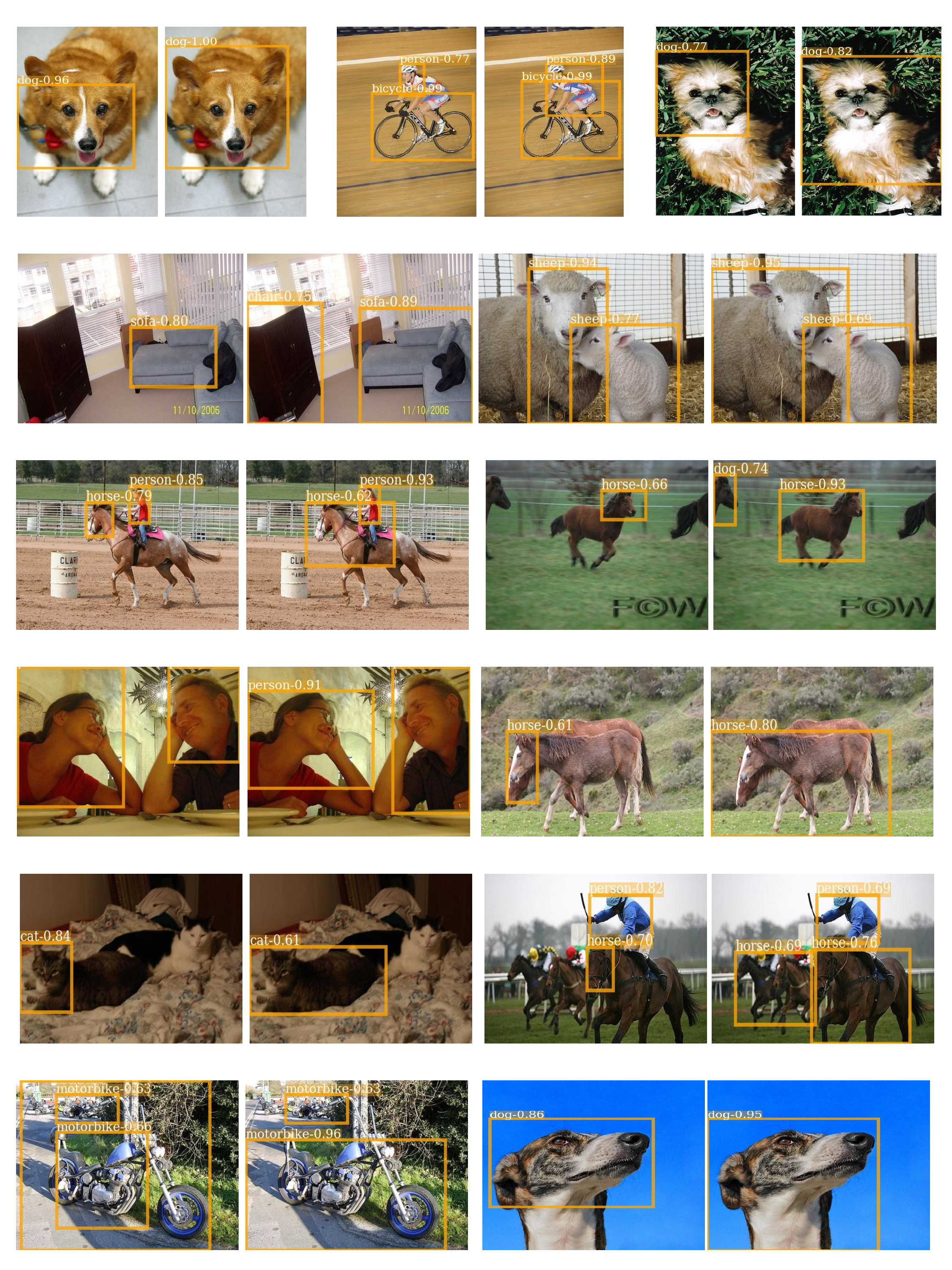}
\caption{Examples that highlight cases of `Part Domination'. For every pair: baseline (left) and our model (right).}
\label{fig:quali_more_part}
\end{figure*}

We show additional results that highlight cases of \textit{`Instance Ambiguity'} and \textit{`Part Domination'} in Fig.~\ref{fig:quali_more_inst} and Fig.~\ref{fig:quali_more_part}, respectively. Following the main paper, we compare our final model to a baseline without the modules proposed in Sec.~\ref{sec:refine} and Sec.~\ref{sec:dropout} of the main paper to demonstrate the effectiveness of these two modules visually.  We show a set of two pictures side by side, the baseline on the left and ours on the right. From the results, we observe: (1) we have addressed the \textit{`Missing Instances'} issue and previously ignored objects are detected with great recall (\eg, monitor, sheep, car, and person in Fig.~\ref{fig:quali_more_inst}); (2) we have addressed the \textit{`Grouped Instances'} issue as our model predicts tight and precise boxes for multiple instances rather than one big one (\eg, bus, motor, boat, car in Fig.~\ref{fig:quali_more_inst}); (3) we have also alleviated the \textit{`Part Domination'} issue for  objects like dog, cat, sheep, person, horse, and sofa (see Fig.~\ref{fig:quali_more_part}).

\begin{figure*}[t]
\centering
\includegraphics[width=0.98\textwidth]{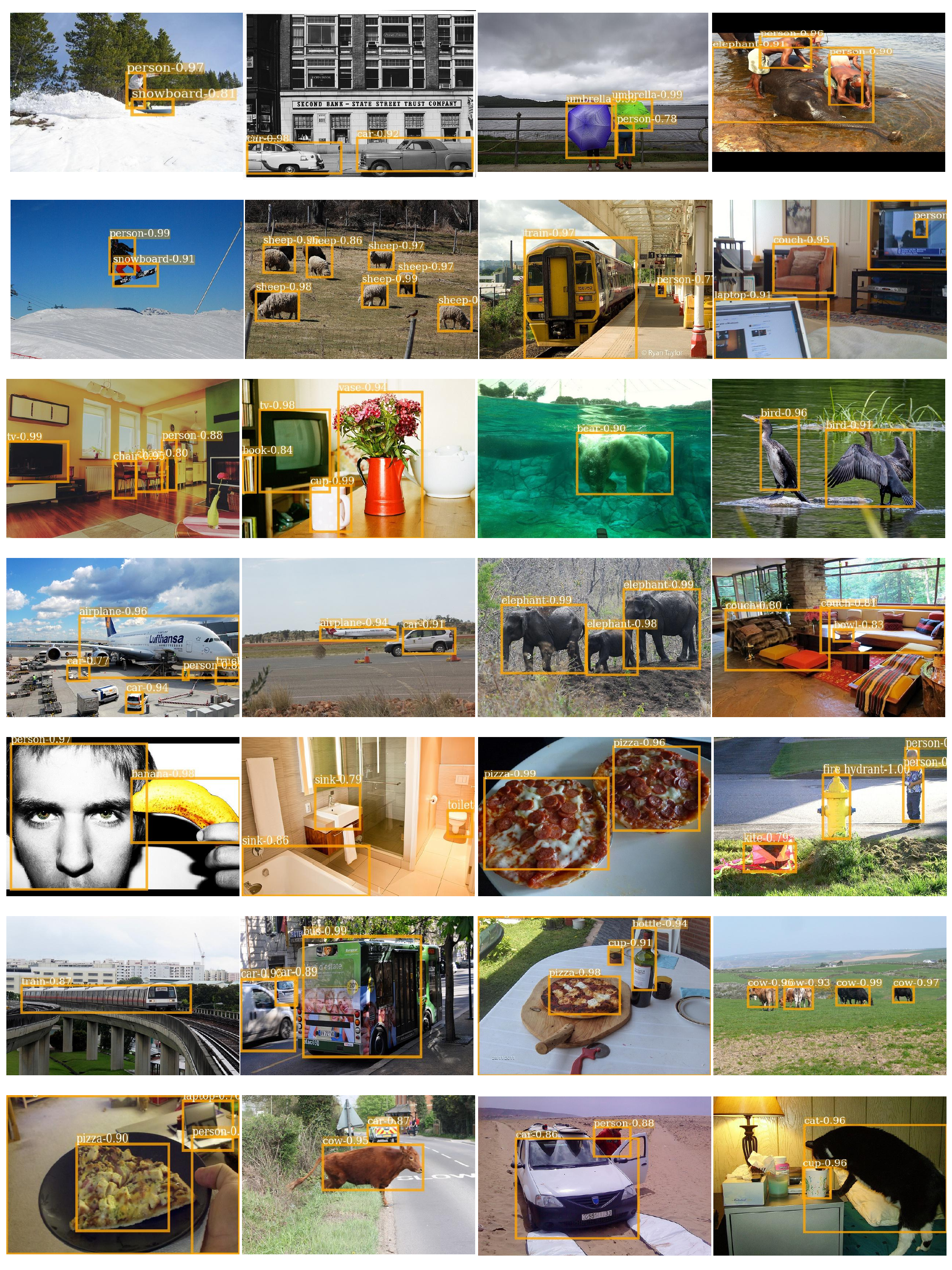}
\caption{Additional visualization results of the proposed method on the COCO2014 validation set.}
\label{fig:quali_more_coco}
\end{figure*}

We also provide additional visualization of our results on COCO in Fig.~\ref{fig:quali_more_coco}. We obtain these results by running the VGG16 based model on the COCO 2014 validation set. Our model is able to detect different instances of the same category (\eg, car, elephant, pizza, cow, umbrella) and various objects of different classes in relatively complicated scenes, and the obtained boxes can cover the whole objects pretty well rather than simply focusing on discriminative parts.

\subsection{Results on ImageNet VID dataset}
\begin{figure*}[t]
\centering
\includegraphics[width=\textwidth]{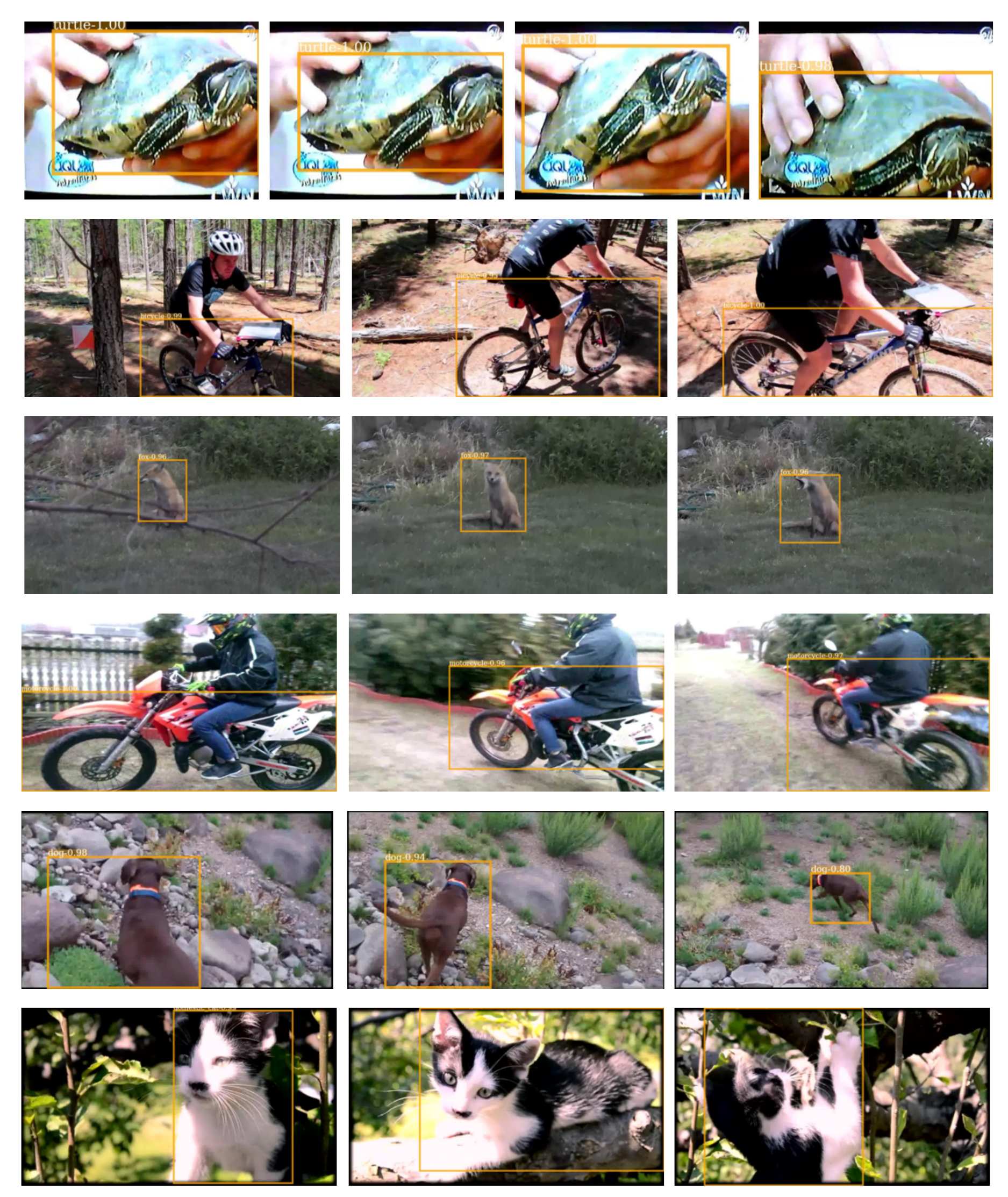}
\caption{Additional visualization results of the proposed method on the ImageNet VID validation set.}
\label{fig:quali_more_vid}
\end{figure*}

Additional visualizations of our obtained results on ImageNet VID are shown in Fig.~\ref{fig:quali_more_vid}, where the frames of the same video are illustrated in the same row. These results are obtained using the ResNet-101 based model. We observe: our model is able to handle objects of different poses, scales, and viewpoints in the videos. 

\section{Proposal statistics}
For consistency with prior literature, we use Selective-Search (SS)~\cite{ss} for VOC and MCG~\cite{mcg} for COCO. Both methods generate around 2K proposals on average as shown in Tab.~\ref{table:prop} but occasionally yield more than 5K on certain images. Our Sequential batch back-propagation can handle these cases easily even with ResNet-101, while other methods quickly run out of memory (Fig.~\ref{fig:mem-exp} in main paper). 

\begin{table}[h]
\centering
\resizebox{\columnwidth}{!}{
\begin{tabular}{c | c c c  c c c }
\specialrule{.15em}{.05em}{.05em}
Data & voc07-train & voc07-val & voc07-test & voc12-train & voc12-val & voc12-test  \\
Avg/Max & 2001 / 4663 & 2001 / 5236 &  2002 / 5398  & 2014 / 5254  & 2010 / 5563  & 2020/5660 \\
\hline
Data & coco14-train & coco14-val & coco17-train & coco17-val & coco-test   & - \\
Avg/Max & 1957 / 5143 &  1958 / 6234  & 1957 / 6234 & 1961 / 3774  & 1947 / 4411  &-  \\
\specialrule{.15em}{.05em}{.05em}
\end{tabular}}
\caption{Proposals statistics.}\label{table:prop}
\end{table}

\section{Need for redundant proposals} 
\label{app:prop}
In WSOD, since ground-truth boxes are missing, object proposals have to be redundant for high recall rates, consuming significant amounts of memory. To study the need for a large number of proposals we randomly sample $p$ percent of all  proposals. A VGG16 based model on VOC 2007 is used. The results are summarized in Tab.~\ref{table:prop-exp}. Reducing the number of proposals even by a small amount significantly  reduces accuracy: using 95\% of the proposals causes a 2.8\% AP drop. This suggests that all proposals should be used for  best performance. 

\begin{table}[h]
\centering
\small{
\begin{tabular}{c | c c  c c |c }
\specialrule{.15em}{.05em}{.05em}
$p$  & 60\% & 80\% & 90\%  & 95\% & 100\% \\
\hline
AP  & 48.4  & 49.7 & 50.8 & 52.1 & 54.9 \\
\specialrule{.15em}{.05em}{.05em}
\end{tabular}}
\caption{Effect of using different number of proposals.}
\vspace{-1em}
\label{table:prop-exp}
\end{table}

\section{Additional details on video experiments}
\label{app:vid}
In this section, we provide additional details of Sec.~\ref{sec:video}. Following supervised methods for video object detection~\cite{zhu17fgfa, xiao-eccv2018}, we experiment on the most popular dataset: ImageNet VID~\cite{imagenet}. Frame-level category labels are available during training. For each video, we use the uniformly sampled 15 key-frames from~\cite{zhu17fgfa} for training. For evaluation, we test on the standard validation set, where per-frame spatial object detection results are evaluated for all the videos.

The two models `Ours' and `Ours (MIST only)' are two single-frame baselines with or without Concrete DropBlock (main paper Sec.~\ref{sec:dropout}). In addition, the memory-efficient sequential batch back-propagation (main paper Sec.~\ref{sec:step-bp}) permits to leverage  short-term motion patterns (\ie, optical-flow) to further increase the performance. For `Ours+flow,' we first use FlowNet2~\cite{flownet2} to compute optical flow between  neighboring frames and the reference frame. The estimated flow maps are then used to warp the nearby frames' feature maps to linearly sum with the reference frame for representation enhancement. The accumulated features are then fed into the proposed task head (modules after `Base' in main paper Fig.~\ref{fig:head}) for weakly supervised training. This method combines the flow-guided feature warping method as discussed in~\cite{zhu17fgfa} to leverage temporal coherence and the proposed WSOD task head to handle frame-level weak supervision. Hence it achieves better results than the aforementioned two baselines (`Ours' and `Ours (MIST only)') using both VGG16 and ResNet-101 as reported in Tab.~\ref{table:vid}.

\end{document}